\documentclass[lettersize,journal]{IEEEtran}
\usepackage{amsmath,amsfonts}
\usepackage{algorithmic}
\usepackage{algorithm}
\usepackage{array}
\usepackage[caption=false,font=normalsize,labelfont=sf,textfont=sf]{subfig}
\usepackage{textcomp}
\usepackage{stfloats}
\usepackage{url}
\usepackage{wrapfig}
\usepackage[table]{xcolor}
\usepackage{color}
\usepackage{verbatim}
\usepackage{subcaption}
\usepackage{graphicx}
\usepackage{cite}
\usepackage{hyperref}

\begin{document}

\title{Self-supervised cross-modality learning for uncertainty-aware object detection and recognition in applications which lack pre-labelled training data}

\author{Irum Mehboob, Li Sun, Alireza Astegarpanah and Rustam Stolkin
\thanks{Irum Mehboob, Alireza  Rastegarpanah and Rustam Stolkin are with the  Extreme Robotics Lab, School of Metallurgy and Materials, The University of Birmingham, Birmingham, B15 2TT, UK. (email: ixm602@student.bham.ac.uk), (e-mail: a.rastegarpanah@bham.ac.uk)and (email: r.stolkin@bham.ac.uk) }
\thanks{Li Sun is with Horizon Robotics (Motor Vehicle Manufacturing), Beijing, China.
China, and also with The University of Sheffield, S10 2TN Sheffield, U.K.
(e-mail: li.sun@sheffield.ac.uk) }}

\maketitle

\begin{abstract}
This paper shows how an uncertainty-aware, deep neural network can be trained to detect, recognise and localise objects in 2D RGB images, in applications lacking annotated training datasets. We propose a self-supervising ``teacher-student'' pipeline, in which a relatively simple “teacher” classifier, trained with only a few labelled 2D thumbnails, automatically processes a larger body of unlabelled RGB-D data to teach a “student” network based on a modified YOLOv3 architecture.
Firstly, 3D object detection with back projection is used to automatically extract and ``teach'' 2D detection and localisation information to the student network. Secondly, a weakly supervised 2D thumbnail classifier, with minimal training on a small number of hand-labelled images, is used to teach object category recognition. Thirdly, we use a Gaussian Process (GP) to encode and teach a robust uncertainty estimation functionality, so that the student can output confidence scores with each categorization.
The resulting student significantly outperforms the same YOLO architecture trained directly on the same amount of labelled data. Our GP-based approach yields robust and meaningful uncertainty estimations for complex industrial object classifications. The end-to-end network is also capable of real-time processing, needed for robotics applications.
Our method can be applied to many important industrial tasks, where labelled data-sets are typically unavailable. In this paper, we demonstrate an example of detection, localisation, and object category recognition of nuclear mixed-waste materials in highly cluttered and unstructured scenes. This is critical for robotic sorting and handling of legacy nuclear waste, which poses complex environmental remediation challenges in many nuclearised nations.
\end{abstract}


\section{Introduction}
\subsection{Motivation}
\IEEEPARstart{T}{his} paper addresses the computer vision problems of detecting, recognising and localising objects. Our proposed method has broad potential to be used for many applications and object types. It is especially useful for industrial or applied problems, where large amounts of application-specific annotated training data are typically unavailable. We demonstrate such an application with a motivating example of robotics challenges in extreme environments, for example, robotic sorting of nuclear waste objects and materials, for the safe remediation of legacy nuclear facilities \cite{OECD2022}.

The UK alone contains an estimated 4.9 million tonnes of legacy nuclear waste \cite{NDA2013}, much of it dating back many decades. Waste items can comprise numerous objects, e.g. contaminated gloves, respirators, swabs, tools, containers, and pipework sections. At the Sellafield site (dating back to the 1940s), a new plant is being built which will use robot arms for the next 50 years. These will cut open old containers, for which there is some uncertainty about the contents. The robots must sort and identify waste items, separate them according to the estimated hazard level, and repackage them into safer modern containers. In addition to the potential for computer vision to help guide robots during e.g. pick and place operations, there is also a need to create inventory lists for the contents of the new containers. Since the quantities of waste are extremely large, automating such inventory generation will be necessary. It is also an essential requirement to estimate and document the uncertainty associated with the inventory for each storage container.

This, and many other real-world industrial problems, pose particular challenges for modern computer vision approaches. Large, annotated, and ground-truthed data-sets are generally unavailable and may be prohibitively difficult, slow, or expensive to create. For example, it has been estimated that labelling the benchmark ImageNet dataset \cite{ILSVRC15}, with 14 million images, took approximately 22 human years of effort.

Meanwhile, the objects and materials in industrial (or domestic) waste-handling problems are extremely diverse and unstructured, often appearing in arbitrary random heaps. For example, a contaminated rubber glove can appear in numerous different shapes and configurations. To incorporate such a perception system with robots, e.g. for autonomous grasping, relatively fast processing speeds are needed. Furthermore, for optimal robotic action planning \cite{rastegarpanah2021semi, dearden2013approach}, the system needs to make explicit use of representations of uncertainty. 

\subsection{Background}

In recent years, modern computing hardware has enabled rapid advances in computer vision recognition tasks, via deep neural network structures.
However, these methods are predominantly based on extensive supervised learning, depending on very large training data-sets, in which each image must be laboriously hand-labelled with ground-truth information. As a result, much of the deep learning computer vision literature is demonstrated on open-source benchmark data-sets.

Many of these benchmark data-sets feature domestic objects, e.g. furniture, kitchen utensils etc., which do not readily transfer to practical industrial problems. In addition to the labour-intensive nature of collecting and hand-labelling data, such human labour can be prone to error. Sometimes an object may not be accurately bounded by bounding boxes or may be assigned a wrong class label. In some cases, it is difficult for a human annotator to categorize some images \cite{IntroWeaklySL}. Many objects, e.g. a cat with a long tail, or a frying pan with a long handle, do not neatly fit within a bounding box. It is not clear what the correct definition of a bounding box should be, since a complete bounding box will contain large areas of non-object background pixels. Conversely, a box that is tightly fitted to the body of the cat or the frying pan, will omit key parts and features of these objects (the tail or handle). An uncertainty-aware approach to image-based learning, is valuable for such problems.

State-of-the-art CNN based object detectors such as Mask Region-based convolutional neural network (Mask R-CNN) \cite{he2017mask}, Fast Region-based convolutional neural network (Fast R-CNN) \cite{girshick2015fast} and 
Single Shot MultiBox Detector (SSD) \cite{liu2016ssd} have demonstrated impressive object detection capabilities. However, most of these models are unable to estimate the uncertainty accompanying each detection or classification.

More recently, object detection YOLOv3\cite{yolov3} network does assign a confidence estimate alongside its output detections. However, this capability is conventionally trained by inputting confidences that are derived from a simplistic calculation (essentially defining ``confidence'' as the proportion of the network's output box which overlaps with the ground-truth bounding box).  

\begin{figure}[h]
    \centering
    \includegraphics[width=\columnwidth]{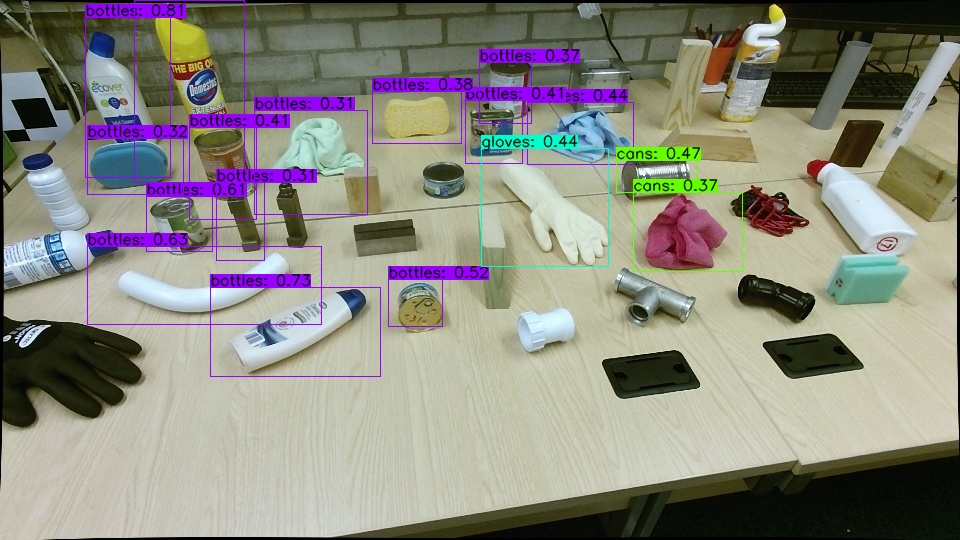}
    \caption{Detections on example image from our nuclear waste test dataset using the standard version of YOLOv3. Note how the conventional YOLOv3 can assign overly high confidence numbers to incorrect classifications.}
    \label{fig:testImgYolo}
\end{figure}

Figure \ref{fig:testImgYolo} shows an image from our dataset \cite{Everingham15}, with detection results from a conventional YOLOv3 \cite{yolov3} network. It can be seen that this model was not successful in recognising the object categories accurately, with most of the objects labelled as ``bottle''. Furthermore, the confidence scores assigned to each detection are questionable, e.g. a ``plastic-pipe'' is detected object as ``bottle'' with a high confidence score of 0.63. 

In this work, we have chosen to use YOLOv3 \cite{shen2023improvedYolov3} over its more recent variants for the following reasons. You Only Look Once (YOLOv3) is a fast object detector that integrates the feature pyramids network and achieves a good balance between detection accuracy and detection speed, making it one of the most popular methods in this field. Redmon and Farhadi \cite{yolov3} proposed a balanced and optimised algorithm regarding the speed and accuracy of object detection.

Later variants of YOLOv3 has been developed such as v4,v5,v6 and v7 \cite{wang2023yolov7}. New variants have developed an efficient backbone and a more understandable label assignment strategy and have minimal to no impact on calculation overhead. Despite that, YOLOv3 is still providing the base network to these variants. It is still very popular in the research community as it provides a simple implementation and deployment structure \cite{di2022multiyolov3}. Ge et al. \cite{ge2021yolox} articulate this perspective by stating that, while YOLOv4 and YOLOv5 have indeed made significant strides in object detection accuracy, they may potentially grapple with issues pertaining to over-optimization. The YOLOv3 algorithm is a popular choice in the industry for its high detection efficiency among the YOLO family, with a broad range of applications in various domains such as

human nail abnormality detection \cite{pellegrino2023effect}, pavement distress detection \cite{du2021pavement}, pedestrian detection \cite{Gongpedestrian}, tracking smart robot car \cite{garciaTrackingrobots} apple growth stage detection \cite{Tian2019apple}, industrial distress detection \cite{Tang_2021surface_detection}, and perception systems for driver-less cars \cite{Gong2022driveless}.

The main motivation for our use of YOLOv3 is that it incorporates functionality for explicitly encoding and outputting an estimate of confidence alongside its object categorization decisions.
\begin{figure*}[h]
    \centering
    \includegraphics[width=7.16in]{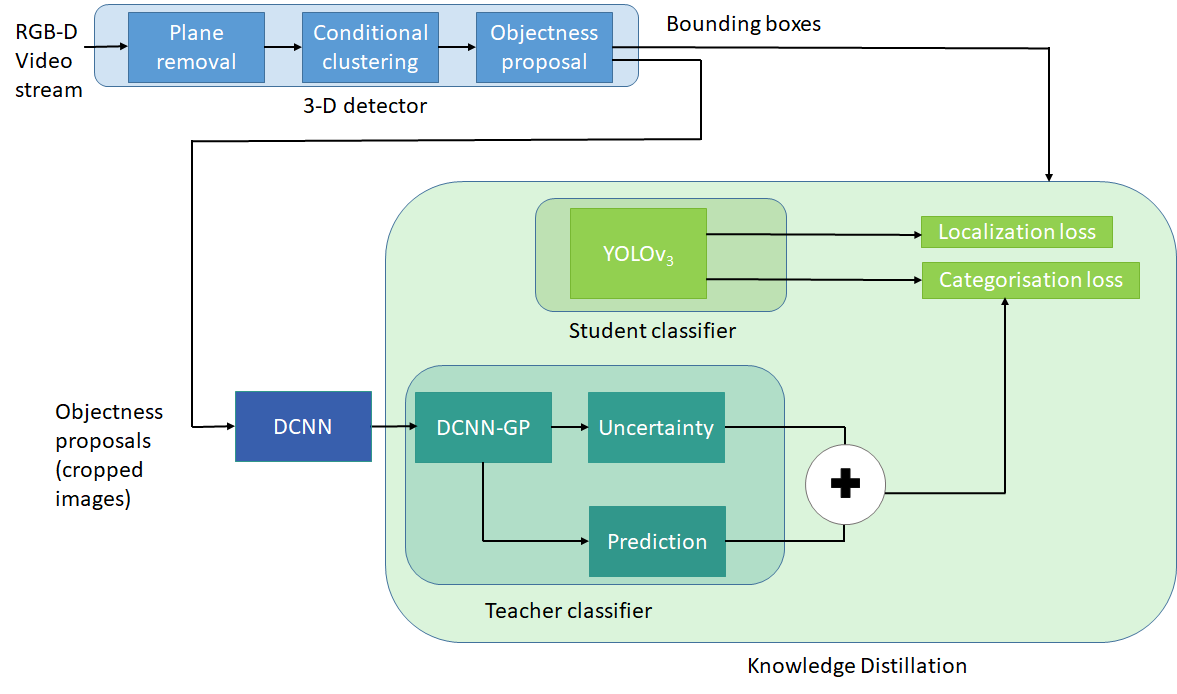}
    \caption{The outline of the proposed method for rapidly boot-strapping a learning system, in a semi-supervised manner, requiring relatively sparse data. This is accomplished by combining Gaussian Processes and YOLOv3 in a Knowledge Distillation paradigm.}
    \label{fig:Architecture} ,
\end{figure*}

Later versions of YOLO do not possess this functionality. In our work, we modify and enhance this uncertainty-awareness functionality by using a Gaussian Process to model uncertainty in a teacher classifier. The teacher then teaches robust uncertainty estimations to our modified YOLO3 classifier during teacher-student training.

\subsection{Approach and novel contributions}
We use our previous work \cite{Kevinwork} as a baseline method, which also introduced our nuclear waste objects computer vision data-set. This method successfully detected objects, and accurately assigned category labels compared to contemporary methods from the literature. However, it was computationally expensive (execution time for detection was 100ms-200ms). It sometimes made false positive detections of background regions as objects, and object category assignment could be noisy and variable. It also struggled to detect small objects or partially occluded objects in cluttered scenes. Most importantly, this system also lacked an ``uncertainty-aware'' functionality.

In this study, we address these problems. We describe a new approach which yields more accurate detections, with less computational complexity, while adding a new functionality enabling the system to output confidence estimates to accompany each detection.

Common sense suggests that a robust model, with a meaningful and useful ``uncertainty-awareness'' capability, should output low confidence scores whenever it outputs false-positive detections or incorrect object category labels. In contrast, as seen in Fig. \ref{fig:testImgYolo}, in our example nuclear waste application we can see that the conventional approach to training confidence estimates in YOLOv3, often results in inappropriate output confidence scores during testing. To provide an improved uncertainty-awareness capability, this study proposes the fusion of a Gaussian Process (GPC) model for classification with a YOLOv3 detector, in a ``teacher-student'' paradigm, enabling real-time detection accompanied by robust and useful confidence scores. In contrast to previous methods for assigning confidences (discussed above), we adopt a ``teacher-student'' approach (related to ``knowledge distillation'' methods \cite{hinton2015distilling}. We use the GPC as the teacher and YOLOv3 as the student. The GPC proposes confidence scores associated with object image thumbnails and teaches these confidences to the YOLOV3 network during its object category recognition training.

\begin{figure*}[h]           
    \centering
    \includegraphics[width=7.16in]{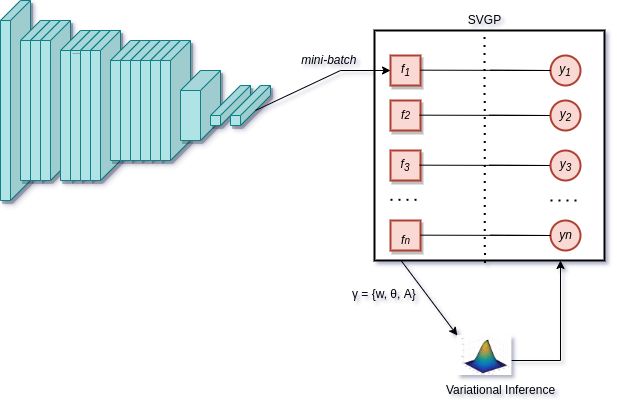}
    \caption{Deep kernel learning architecture with Stochastic variational inference procedure.}
    \label{fig:DCNN-GPC}
\end{figure*}

Note that previously, ``knowledge distillation'' has been used in a very different way. Typically a complex (and computationally expensive) strong classifier is used as the ``teacher'', and trains a simpler (and cheaper) classifier which serves as the ``student'' \cite{goldt2019dynamics}. I.e. a large amount of knowledge, encoded in the large and complex teacher network, is ``distilled'' \cite{abbasi2020modeling} into a much smaller and computationally cheaper student network.

In contrast, a key novelty of our work is that we show how a relatively simple and cheap classifier can be bootstrapped as a ``teacher'', which generates inputs to a much more complex and powerful ``student'' classifier during its training. The resulting strong classifier (student) then outperforms its teacher and also outperforms the same network structure when trained in a conventional way, without the teacher, on the same data-set. 

First, we use a 3D-detector from our previous work \cite{Kevinwork} to generate objectness proposals from RGB-D video streams, and generate corresponding 2D object thumbnails from the RGB-D data. We manually label a small number of these thumbnails. Some are retained for testing, and a few are used as a training input to a``weakly supervised'' system. The system then bootstraps on this small input data, becoming ``self-supervised'. I.e. based on this small labelled data, our system effectively creates and labels more training data, while training itself by using the teacher-student paradigm.

We train the classic pre-trained Resnet-50v2 on this small labelled dataset, by using transfer learning \cite{pan2009survey}. Then we augment this Resnet network with a Gaussian Process (GP) model to provide a sophisticated functionality for learning uncertainty-awareness. The resulting ``teacher'' then generates a much larger scale of automatically labelled, or ``self-labelled'' data as inputs to the training of the YOLOV3 network. Meanwhile, the GP component of the teacher is used to provide input to the uncertainty-awareness learning component of our modified YOLOv3 network (in contrast to the more simplistic uncertainty learning approach of the original YOLOv3 as discussed above).

The knowledge of the ``teacher'' network is thus ``distilled'' into a YOLOv3 ``student'' network, using the variation of loss for classification. This variation of loss is composed of knowledge distillation loss and the sum of squared loss. This technique improves the classification loss compared to the original YOLOV3 object detector method. The resulting network, informed by the GP component of the teacher during training, also generates significantly improved confidence/uncertainty values for each classification, compared to the original YOLOv3.

The main contributions of this paper are as follows:

 \begin{figure*}[h]
    \centering
    \includegraphics[width=7.16in]{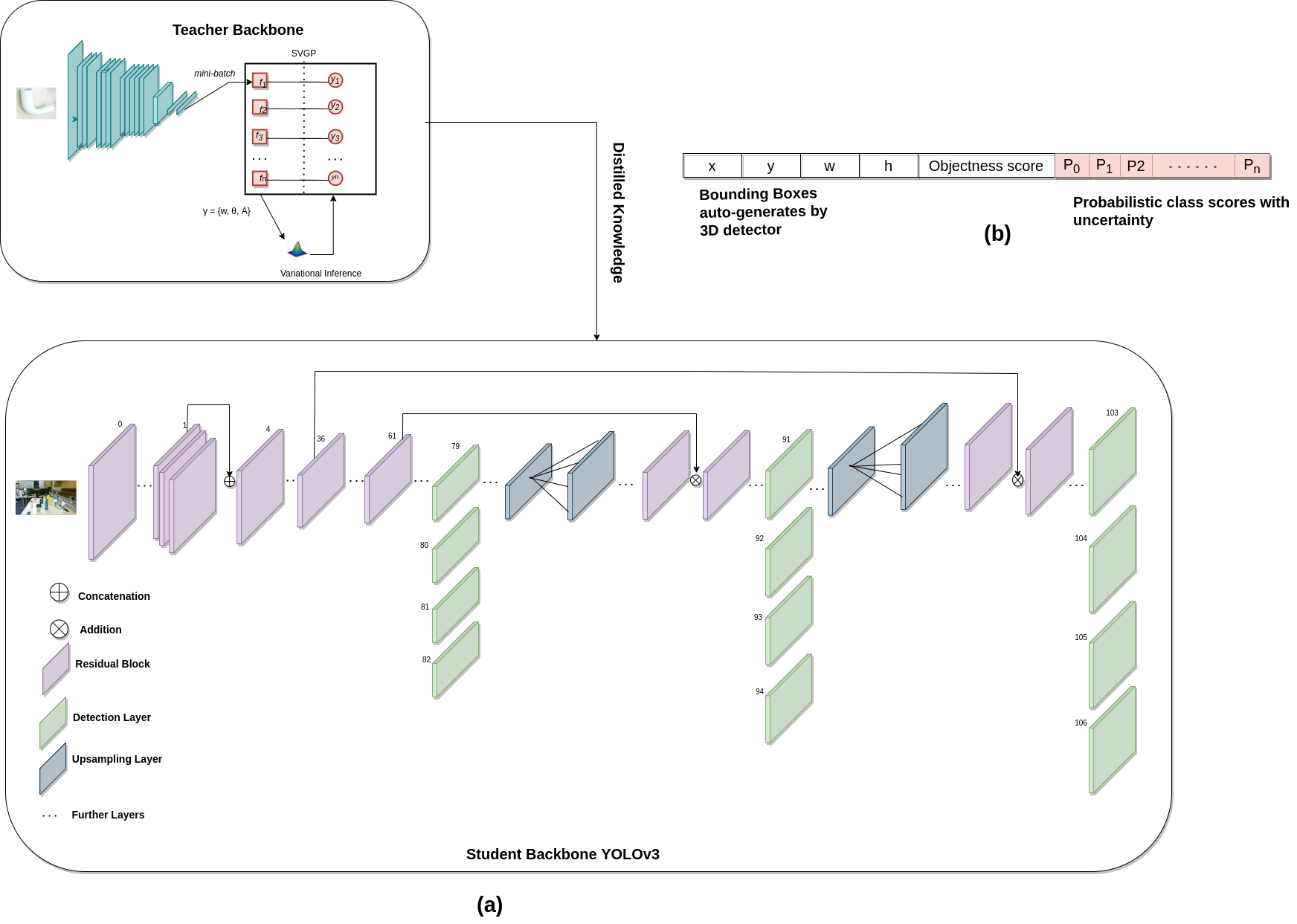}
    \caption{Schematic of the Knowledge distillation pipeline for categorization. a) The transfer of knowledge from the teacher backbone, as shown in Figure \ref{fig:DCNN-GPC}, to the student backbone utilizing the YOLOv3 architecture. (b) Illustration of the YOLOv3 output structure, where bounding box coordinates are generated by a 3D detector, defining the spatial location and size of each detected object within the 3D space. The objectness score indicates the confidence level that the bounding box contains an object. The final part of the output comprises probabilistic class scores, which provide a probabilistic distribution over possible classes, thereby incorporating uncertainty in the classification process.
 }
    \label{fig:my_yologp}
\end{figure*}

\begin{enumerate}
 \item A self-supervised 2D objectness detection, trained by automatically extracting and labelling 2D RGB object thumbnails from 3D RGB-D data. We use 3D conditional clustering within the point clouds to automate the extraction and labelling of bounding boxes without human effort. This automatically generates 2D object bounding box annotations as inputs for training the YOLOV3 network, hence our term ``self-supervised'' learning.
 \item A novel use of ``teacher-student'' and ``knowledge distillation'' concepts, to enable boot-strapping a weak classifier (based on a small amount of annotated training data) to train a more complex and strong classifier (by automatically generating and feeding it training examples).

 Not that this is significantly novel in contrast to conventional knowledge distillation methods. Such methods use complex, strong classifiers, to teach effective classification capabilities to a smaller, simpler classifier (e.g. for implementations on small processors). In contrast, we show how to invert this concept, using a weakly trained classifier to automatically generate large amounts of training data for teaching a larger and more complex student classifier, which eventually outperforms the teacher.
  \item We propose a new way to enable a classifier network to learn uncertainty-awareness, i.e. the ability to output a confidence value alongside each object detection and classification decision. In contrast to the conventional YOLOv3 approach, by using items 1) and 2) we enable self-supervised training of a Gaussian Process Classifier (GPC) as part of the ``teacher'' in our teacher-student paradigm. The purpose of the GPC is to teach confidence/uncertainty scores to the YOLO student network, alongside its learning of objectness detection and object category values during teacher-student training. This yields significantly better quality confidence outputs than conventional approaches to YOLOV3 confidence training, in our example industrial waste objects application.
  \item We redesign the loss function of classic knowledge distillation, which works more effectively with our waste object data-set and achieves SOTA performance, while reducing computational complexity.
  \item Our semi-supervised and self-supervised methods can be readily applied to new industrial applications, where no large ground-truthed or annotated training data-sets exist. We demonstrate this capability by using our unique nuclear waste objects data-set, motivated by the robotics and AI challenges of environmental clean-up and remediation on legacy nuclear sites in hazardous environments,

\end{enumerate}

\section{RELATED WORK}

\begin{figure*}[h]
\centering
\begin{tabular}{c}
\subfloat[]{\includegraphics[width =8.5cm]{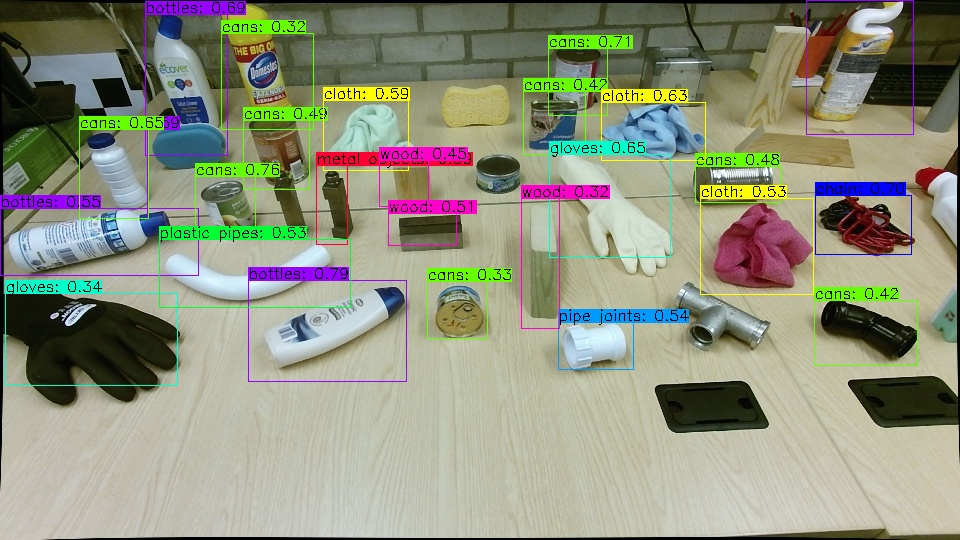}} 
\qquad
\subfloat[]{\includegraphics[width =8.5cm]{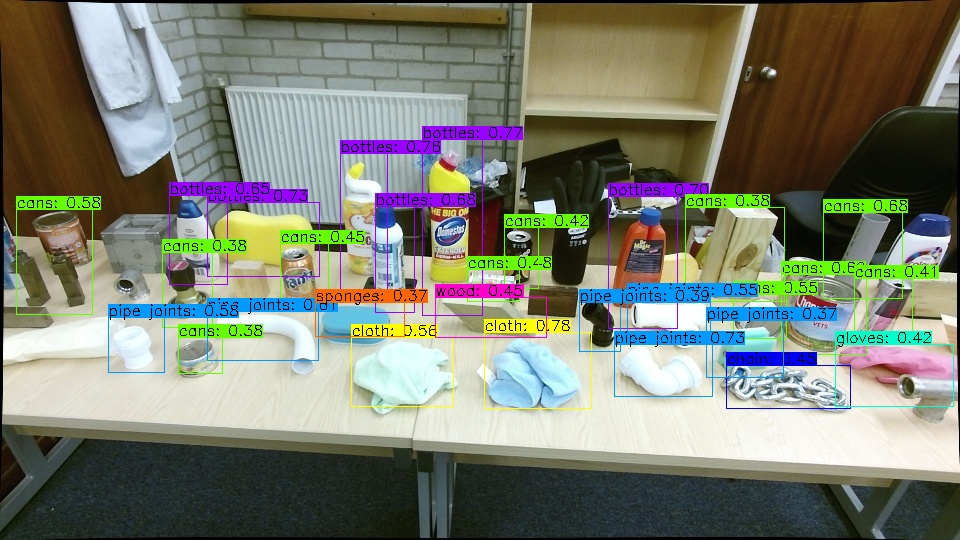}}\\

\subfloat[]{\includegraphics[width = 8.5cm]{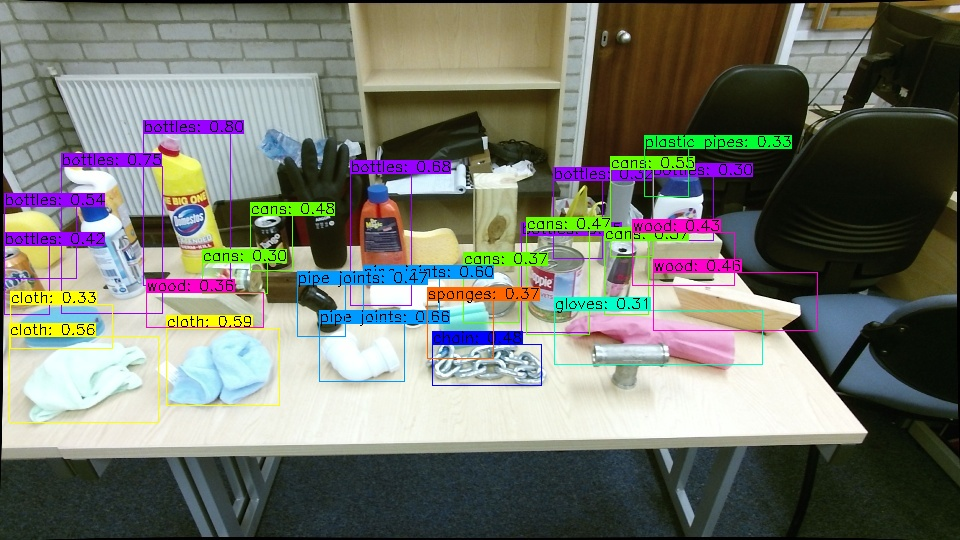}}
\qquad
\subfloat[]{\includegraphics[width = 8.5cm]{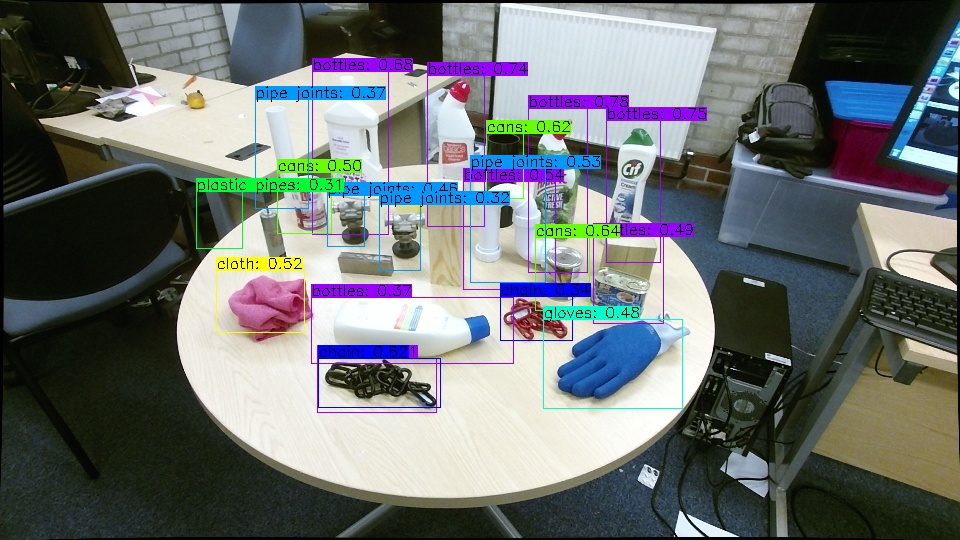}}
\end{tabular}
\caption{Some examples from Nuclear waste test Dataset after training YOLOv3 through Knowledge distillation. As seen in Fig. \ref{fig:testImgYolo} conventional Yolov3 can assign overly high confidence to incorrect classification and does not detect small objects and objects of coarse scale. In contrast, the above Fig. images show the results of YOLOv3 after training through our  "teacher-student" knowledge distillation paradigm. We can see that now YOLOv3 can detect most of the objects and classify them correctly, while also assigning sensible and meaningful confidence scores with each classification. }
\label{fig:testimgesYOLOGPC}
\end{figure*}

In this section, we frame our contributions in the context of related literature on: i) object detection and category recognition in computer vision; ii) weakly-supervised learning; iii) pretext learning; iv) unsupervised and self-supervised learning; v) methods for representing uncertainty; vi) teacher-student and knowledge distillation learning paradigms.

\subsection{object detection and category recognition}
Object detection, and object category recognition, remain challenging problems in modern computer vision research. Such technology is proliferating with the ubiquity and miniaturisation of digital cameras, especially in smartphones. It has numerous applications in areas such as content-based image retrieval \cite{zhou2017recent}, autonomous driving \cite{chen2017multi}, robotic grasping \cite{redmon2015real}, security \cite{jain2019evaluation}, and human-computer interaction \cite{branson2010visual}. 

Early object detection methods required hand-coded, prior knowledge of explicit object properties, such as shape, size, or colour \cite{gennery1992visual,stolkin2008mrf,BALLARD1981Hough, stockman1982matching}. From the 1990s, advances were increasingly made in discovering features that: i) could be ``learned'' or parameterised from a few thumbnail examples, or a single example, of the object in question; and ii) were invariant to changes in size (or range from a camera), orientation (or viewing angle), or motions and configuration changes of deformable or articulated objects such as humans, animals or robots. Key approaches in this era included colour histograms \cite{ennesser1995finding}, Scale-Invariant Features' (SIFT) \cite{lowe1999object}, Histograms of Oriented Gradients (HOG) features \cite{dalal2005histograms}, and Haar features \cite{jones2003fast}. These methods used low-level, ``hand-crafted'' (human-designed) features. They were computationally cheap, and worked well in early commercial applications such as detecting human faces to enable auto-focus in consumer digital cameras \cite{jones2003fast}, or tracking a moving target whose features were learned from the first frame of a sequence \cite{stolkin2008efficient, stolkin2012bayesian}. However, such features were not sufficiently sophisticated to enable generalisable object category recognition, over broad ranges of complex objects.

Since the proliferation of powerful GPU chips, enabling the widespread use of deep neural networks \cite{lecun2015deep}, it has no longer been necessary to devise features, since maximally discriminatory features become automatically (implicitly) encoded into the deep neural network weights during training. Recent methods based on Convolutional Neural Networks (CNN) \cite{FastRNN},\cite{YOLO},\cite{DeepCrack} have greatly advanced object detection and category recognition.

Recent object detection methods can be broadly divided into two types: (i) multi-stage object detection and (ii) one-stage object detection. The multi-stage object detectors, such as R-CNN \cite{girshick2015region}, Fast R-CNN \cite{girshick2015fast} and Spatial Pyramid Posenet (SPPNet) \cite{he2015spatial}, use a ``course-to-fine'' detection process, starting with identifying potential regions of interest, which are refined through successive stages. These multiple steps make these methods relatively computationally expensive and time-consuming to run. However, they tend to perform very robustly in terms of both object localisation and category recognition accuracy. In contrast, one-stage detectors such as YOLO \cite{YOLO} SSD \cite{liu2016ssd} and RetinaNet \cite{lin2017focal} are efficient and fast. However, their accuracy is dependent on very large ground-truthed data-sets such as Pattern Analysis, Statistical Modelling and Computational Learning Visual Object classes (PASCAL VOC) \cite{everingham2010pascal} \cite{everingham2015pascal}, ImageNet \cite{ILSVRC15} and Microsoft Common Objects in Context (MS-COCO) \cite{lin2014microsoft}. I.e. these are ``data hungry'' methods that rely on large amounts of human-annotated data.

While extensive benchmark data-sets exist in the academic community for e.g. domestic household objects or street scenes, such data is rarely available for industrial applications. In our example application of robotic nuclear waste sorting, we need the real-time, or close to real-time speeds offered by YOLO type networks, to support decision-making in industrial robotics applications. However, these specialised industrial applications mean that large, annotated data-sets are not available. We overcome this problem by adopting a combination of semi-supervised and self-supervised approaches.

\subsection{Weakly Supervised Object Detection}
Weakly supervised learning methods offer ways of overcoming the problem of insufficient availability of labelled training data. Three main approaches to weak supervision have been explored in the literature \cite{zhou2018brief}. In ``incomplete supervision'' \cite{settles2009active,hady2013semi}, a subset of training data is given with labels, while a larger part of the training data is unlabelled. In ``inexact supervision'' \cite{foulds2010review}, training data are given only coarse-grained labels; whereas ``inaccurate supervision'', \cite{frenay2013classification,gao2016risk} addresses the problem of poorly labelled data in which the labels are not always correct, with some degree of labelling errors.

In \cite{huang2017learning}, weakly labelled data was used for face detection,  using a subspace-based generative model. In contrast, \cite{ge2020towards} proposed a flexible framework for automatic inspection for industrial object detection. This was an example of incomplete supervision in the sense that not all objects in the image were labelled. This is because it is often difficult and expensive to obtain fully labelled data for industrial images. In \cite{jie2017deep} a method for object detection is proposed using end-to-end CNN models with instance mining algorithms. In \cite{cheng2016semi}, a DCNN was employed as a classifier in co-training of a small-scale labelled data-set.

A drawback of such methods is that they are all parametric in nature and therefore can suffer from over-fitting to the limited training data. To overcome this problem, we incorporate a Gaussian Process for Classification (GPC) model into a weakly supervised learning paradigm. GPC is probabilistic and non-parametric with a Radial Basis Function (RBF) kernel. It does not make any assumptions about the underlying data. Our model is ``incomplete supervision'' in the sense that it learns from a small data-set and is robust to noise, and efficient to train. However, that small data trains a teacher classifier, which then bootstraps with larger additional data that is automatically generated from objectness proposals in a self-supervised learning framework (described below). In our proposed method, the GPC uses features learned from a DCNN. In this way, we achieve accuracy above 80\%, without any over-fitting concerns, on complex industrial waste objects, while also relying on only a very small amount of annotated data. 

\subsection{Pretext learning}
Another way of coping with only a small amount of labelled training data, is to use ``pretext learning''. In pretext methods \cite{misra2020self, kolesnikov2019revisiting} only a small amount of training data exists for the desired application. However, a large amount of annotated data exists for other types of objects. For example, consider a situation in which we wish to develop a system for category recognition of industrial objects. We might not have large annotated industrial object training data, but we do have large public data sets available for domestic household objects. It is possible to ``pre-train'' a network on the easily available object data. This achieves a set of network weights which encode useful knowledge about e.g. ``objectness'' detection (distinguishing groups of object pixels from background pixels), even though the system is not yet attuned to recognising the specific categories of the industrial objects of interest. Those weights are then used as a starting point, for initialisation, upon which a small amount of domain-specific data can be used for further training. This fine-tunes the network, after it is already partially trained for generic object recognition, to recognise objects that are specific to our application. Examples of this approach for practical applications include the previous work of our research team \cite{Kevinwork,zhao2017fully,zhao2020simultaneous}.

Instead of using pretext learning, the method proposed in this paper leverages cross-modality information. We run a model-free (conditional clustering-based) 3D objectness detector on the depth modality (i.e. point clouds) of our original RGB-D data, and this is used to obtain object bounding boxes in the corresponding 2D RGB images. These then act as part of the ``teacher'' system, and transfer the knowledge of ``objectness'' to our end-to-end YOLO 2D object detection/recognition ``student'' network. This avoids the need for pretext data-sets or pretext learning.

\subsection{Unsupervised versus Self-Supervised Learning}

To avoid annotating large data-sets for object detection tasks, self-supervised learning has been used in computer vision and machine learning \cite{masci2011stacked,ranzato2007unsupervised,salakhutdinov2010efficient,vincent2008extracting}. Self-supervised learning learns features from an unlabelled data-set, whether it is in the form of images or videos. This alternative approach does not require human labour for annotation. It can be achieved by a variety of methods such as clustering \cite{bojanowski2017unsupervised,caron2018deep,ji2019invariant}, Generative Adversarial Network (GAN) \cite{donahue2016adversarial,mescheder2017adversarial}, pretext tasks \cite{doersch2015unsupervised,noroozi2016unsupervised,wang2015unsupervised} etc.

Representation learning has also been accomplished using clustering. For example in \cite{yang2016joint}, a recurrent framework is proposed for clustering and optimization of triplet losses for joint representation learning and clustering. In \cite{clusteringLandMapping}, a clustering-based self-supervised learning method is proposed, which pre-trains the model for few-shot segmentation for mapping land-cover. These methods

use ConvNet (convolution neural network) for initial pretext learning, followed by clustering. 
In contrast to general-purpose representation learning models, which are often characterized by their complexity and lack of interpretability, our system employs a conditional clustering approach that obviates the need for pretext tasks. This methodological choice confers several advantages, including a more efficient computational performance, particularly in terms of speed. Whereas representation learning models often require a substantial initial investment of time to capture a wide range of features, our system is optimized for real-time applications, resulting in superior execution speed. Furthermore, the highly optimized nature of our approach yields a level of accuracy that exceeds that of more generalized models. Consequently, for tasks demanding real-time responsiveness and high accuracy, our conditional clustering method represents a more efficacious solution.

In our proposed system, the whole process is self-supervised, except for a very small number of human-annotated thumbnail examples which are used to bootstrap the object recognition part of the ``teacher'' structure. We use a novel approach to automatically generate 2D RGB training data from RGB-D image sequences. More specifically, we use a model-free object detector on the point-cloud depth data modality, then apply conditional clustering to obtain objectness proposals. These are used to automatically generate 2D bounding boxes for objects in the corresponding 3D RGB images. These objects are then automatically labelled by our GPC ``teacher'' with both object category labels, and also uncertainty/confidence values. These then feed into the learning inputs of our larger-scale YOLOv3 end-to-end ``student'' network.

\subsection{Uncertainty Awareness in Object Detection}

Many pattern recognition approaches, including most neural network methods, are able to make classification or detection decisions, but typically cannot simultaneously output an indication of confidence or uncertainty associated with each such decision.

More recently, neural networks have been used with Bayesian modelling to perform prediction while modelling the uncertainty of each decision. However, such work has predominantly focused on uncertainty associated with localisation. Jian et al. \cite{confidence2018ODetection} used the intersection-over-union (IoU), between classifier output bounding boxes and ground-truth 2D bounding box regions, as a simple metric of uncertainty in object localisation. Choi et al. \cite{choi2019gaussian} enhanced this by incorporating Gaussian modelling and loss reconstruction in YOLOv3 to predict localization uncertainty. \cite{GMmodelYolo} propose a combination of experts method. They combine the classification results from a Gaussian Mixture Model (GMM) and YOLO, weighted according to the confidence metric output by each model.

The above methods \textit{localize} the objects while also encoding awareness of localization uncertainties. However, they lack the ability to also \textit{categorize} the objects with \textit{uncertainty estimates of the categorization decisions}. For example, it may be the case that a detected object is not a pedestrian but a cyclist. Hence, an ability to estimate uncertainty in a categorization decision is also very important, in addition to localization uncertainty. This is even more critical on problems where smaller amounts of training data may be available, and where critical decision-making needs to be made based on vision system outputs, e.g. action planning of robots in uncertain or unstructured environments, especially in high-consequence safety-critical industries.

This paper proposes a solution to this challenge. We use an end-to-end detection and categorization pipeline based on a YOLOv3 network. However, during training, we use a Gaussian Process as ``teacher'', to predict categorisation uncertainty probabilities, which become part of the training data fed into (i.e. ``taught'' to) the ``student'' YOLOv3 network. The resulting network is able to output uncertainty estimates alongside each object categorization decision.

\subsection{Knowledge Distillation in Object Detection}

Knowledge distillation has received significant attention in the research of deep learning. The main idea is to utilize computationally expensive, large-scale deep models in real-time applications, by ``distilling'' the knowledge of the large-scale network into a smaller network that is computationally cheap and fast at run-time. \cite{urner2011access} distil the knowledge from a fully supervised teacher model to a student model using unlabeled data in a semi-supervised learning task. The term knowledge distillation is specified to this technique of learning in the seminal paper of \cite{hinton2015distilling}. Different knowledge distillation methods and terminologies have been used in research such as teacher-student learning \cite{hinton2015distilling}, mutual learning \cite{zhang2018deep}, assistant teaching \cite{mirzadeh2020improved}, lifelong learning \cite{zhai2019lifelong} and self-learning \cite{yuan2019revisit}. \cite{wang2018dataset} apply knowledge distillation for compressing the training data, to reduce the training loads of deep learning models.

These methods are predominantly used to distil knowledge from a large network to a small network to compress the model. In contrast, our approach is to boot-strap a small ``teacher'' classifier, trained on minimal data, by using it to train a much larger and stronger classifier based on a deep YOLOv3 architecture. Furthermore, we combine the knowledge distillation philosophies of both teacher-student architecture and also data-set distillation. In this paradigm, we use a GPC model as teacher and train it on a small subset of data. We then deploy this model to automatically generate class probabilistic scores based on weakly supervised learning. It distills the knowledge of features and data to the large YOLOv3 model which serves as the ``student''.

%%%%%%%%%%%%%%%%%%%%%%%%%%%%%%%%%%%%%%%%%%%%%%%%%%%%%%%%%%%%%%%%%%%%%%%%%%%%%%%%%
%%%%%%%%%%%%%%%%%%%%%%%%%%%%%%%%%%%%%%%%%%%%%%%%%%%%%%%%%%%%%%%%%%%%%%%%%%%%%%%%%

\section{METHODOLOGY}
Our method comprises the following steps: i) high-quality objectness proposals are generated in RGB-D video streams 

by building on our real-time 3D-based object detection methods of \cite{Kevinwork}; ii) probabilistic object classification of 3D thumbnails using a GPC model which is weakly-trained on a small number of hand-labelled images;
iii) end-to-end training of a YOLOv3 detector through self-supervised learning and a teacher-student paradigm, in which the GPC model teaches object classification and confidence values to the YOLOv3 network, while it trains on a large number of object instances that are automatically generated by the method of i).

\subsection{3D Objectness Detection and Automatic Generation of Large-scale 2D Data}\label{Method_3DObjectness_Detection}
Our objectness detection approach extends methods we developed in \cite{Kevinwork} for real-time 3D objectness detection. Salient region proposals are obtained through point cloud segmentation. RANSAC is used to detect and remove large planes, so that we can obtain only the table-top and ground-top objects. Additionally, we incorporated a more efficient conditional clustering approach to acquire objectness proposals, using perceptual grouping \cite{UnsupervisedSegmentation} based on colour, shape and spatial cues.
We first detect large planes (using RANSAC) in point clouds and remove them, as we are interested in table-top or ground-top objects. We then use a multi-cues conditional clustering approach based on colour, shape and spatial cues to acquire objectness proposals. Given two voxels $p_1$ and $p_2$, the connectability between them $\mathcal{C}(p_1,p_2)$ is defined by distance connectability $\mathcal{C}_d(p_1,p_2)$, color connectability $\mathcal{C}_c(p_1,p_2)$ and shape connectability $\mathcal{C}_s(p_1,p_2)$:

\begin{equation}
    C_s(p_1, p_2) =    
     \max \left( 0, \min \left( 1, \sigma_s - \left( \frac{n_{p_1} \cdot n_{p_2}}{\| n_{p_1} \| \| n_{p_2} \|} \right) \right) \right)\\    
\end{equation}

\[C_c(p_1, p_2) = \max \left( 0, \min \left( 1, \sigma_c - \left\| I_{p_1} - I_{p_2} \right\| \right) \right)\\\]

\[C_d(p_1, p_2) = \max \left( 0, \min \left( 1, \sigma_d - \left\| p_1 - p_2 \right\| \right) \right)
\]

\[C(p_1, p_2) = C_d(p_1, p_2) \cap \left( C_s(p_1, p_2) \cup C_c(p_1, p_2) \right)
\]

where $\textbf{n}_{p1},\textbf{n}_{p2}$ are the surface normals, ${I}_{p1},{I}_{p2}$ refer to the intensity values of $p_1,p_2$, and $\sigma_d$,   $\sigma_c$,  $\sigma_s$ are the connectability thresholds. Neighbouring voxels are then clustered iteratively through this connectability criteria until all clusters become constant.  Parameter values $\sigma_d$,   $\sigma_c$,  $\sigma_s$ are set as 2.0$cm$, 8.0 and 10$^{\circ}$, perform well for our application.

Given 3D objectness proposals detected in 3D world coordinates, each point in the proposal $p(x_w,y_w,z_w)$ can be back-projected to its 2D image coordinates $(u,v)$ and it depth $d$: 
\begin{equation}\label{eq:2d3d}
d~ 
\begin{bmatrix}
    u\\
    v\\
    1
\end{bmatrix} =
C~
\begin{bmatrix}
    R~t\\
    0~1\\
\end{bmatrix}
~
\begin{bmatrix}
    x_w\\
    y_w\\
    z_w
\end{bmatrix}
\end{equation}

where $C$ is the camera intrinsic matrix, and $R$ and $t$ are the rotation matrix and transformation vector respectively. A 2D bounding box is formed for each 3D objectness proposal, which is then used as training data. YOLOv3 takes this data as input, and learns to detect and localise object regions in 2D RGB images, as explained in the following section.

As a result, 2D bounding boxes with boundary-aware segmentation is achieved from 3D objectness proposals. Our method thus automatically generates a large-scale \textit{partially annotated} dataset. This comprises: i) a large set of 2D images, in which every image has been automatically annotated with ground-truth bounding boxes for all objects; ii) cropped object thumbnail images for each bounding box. This partially annotated data \textit{does not yet have object category annotations}.

\subsection{Probabilistic Object Classification for the Teacher Classifier}\label{Method_ObjectCategoryClassification}
We use a weakly-supervised method to train a DCNN architecture to form part of the ``teacher'' process that enables our self-learning pipeline. Firstly, we create the dataset, as described above, using the 3D detector, Fig.\ref{fig:Architecture}. Secondly, we manually label the object categories for a small subset of 2D object thumbnail images from the automatically generated large-scale dataset. Thirdly, the DCNN (ResNet50) \cite{DeepRL}, which is pre-trained on public large-scale dataset ImageNet, is subjected to further training on this small set of cropped and labelled image data, i.e. a transfer-learning paradigm. Finally, the DCNN-GPC is trained to predict the probabilities (confidence scores) for each category of the cropped images.

The resulting DCNN-GP classifier can now be used to: i) automatically label the object categories of all remaining images and thumbnails of the large-scale dataset; ii) automatically predict probabilities (confidence levels) for each such classification. The weakly-trained DCNN-GP can thus be boot-strapped to strongly train (i.e. ``teach'') the end-to-end learning of a large YOLOv3 network. The Architecture of the teacher network is described as follows.

Our previous work on this topic, \cite{Kevinwork}, used VGG-16\cite{simonyan2014deep} for object classification. In this paper, we use ResNet50, since it is a more recent and efficient model. This network is similar to VGG-16 except for the new identity mapping capability. These features enable ResNet to overcome the problem of vanishing gradients. This is important, because our nuclear waste object dataset is very challenging, and it contains many objects which can be easily confused with each other, e.g. separate categories of plastic pipes and pipe joints. The dataset also includes objects with shiny, reflective surfaces (e.g. metal objects of various kinds), and a variety of deformable objects (e.g. rubber gloves, chains, hoses, sponges, face-masks). These objects are significantly different from those in datasets more commonly used in mainstream academic research on object detection \cite{li2020object, rehman2020deep, kundalia2020multi}.\\

We added two fully connected layers with 4096 hidden nodes at the end of the network to avoid over-fitting and removed the softmax layer. This helps in increasing the accuracy. We set these parameters according to experience gained in our previous related work. Our network is pre-trained on ImageNet \cite{ILSVRC15}. 

During this stage, the features from the last layers \begin{math}(f_{c1} \epsilon \mathbb{R}^{4096} ) \end{math} and \begin{math}(f_{c2 }\epsilon \mathbb{R}^{4096} ) \end{math} are given as input \textit{X} \begin{math}( \epsilon \mathbb{R}^{8192} ) \end{math} to the GPC model. 

The GPC is trained using the small set of hand-labelled object thumbnail data described above. We use a multi-class classification GP model. Essentially this is a regression problem from an input \begin{math} \textit{x} \epsilon \textit{X} \end{math} to discrete labels \begin{math}\textit{y} \epsilon \textit{Y}
\end{math}.
We adopted the Wilson et al. method \cite{wilson2016stochastic} as the basis for our experimental design, with this critical modification to the approach by incorporating features from transfer learning into the GPC model. This adjustment is predicated on the hypothesis that utilizing pre-trained models to extract features can enhance the model's ability to capture relevant information from the data, thereby potentially improving the overall predictive performance of the GP framework.
Rather than using the whole dataset as input, we use a subset of dataset \(Z \in \mathbb{R}^{m \times D}\)
 known as inducing points where $m \subset$ total number of observations \textit{n}, and \textit{D} is the input dimensionality. We use inducing points to overcome computational complexity  \(O(n^3)\).
For these inducing points we define latent variables $u_j$ and employ a variational distribution \textit{q(u)} to approximate the true posterior of \textit{u}.

The optimization of the variational distribution and the inducing points is achieved by maximizing the evidence lower bound(ELBO) on the log marginal likelihood. 

\begin{equation}
\log p(y) = \int q(u) \mathbb{E}_p(f | u) [\log p(y | f)] dZ - 
KL[q(u) || p(u)]
\end{equation}
in this equation, $KL$ represent the \textit{Kullback-Leibler} divergence between the variational distribution and the prior over $u_j$

To predict the class of a new input $x^*$, the classification model integrates over the variational posterior distribution of the latent variables:

\begin{equation}
    p(y^*| X, Y, x^*) = \int p(y^* | u)q(u)du
\end{equation}

Class probabilities are obtained by applying a softmax function to these integrated outputs, enabling effective classification of $x^*$.
This approach allows the model to produce probabilistic predictions for classification tasks, integrating the strengths of deep learning transformations with the probabilistic modelling capabilities of Gaussian Processes.

We employ a Stochastic Variational Gaussian Process (SVGP) \cite{GPBigData} due to its enhanced scalability and computational efficiency, particularly when handling large datasets that exceed the practical processing capabilities of traditional Gaussian Processes (GPs). This overcomes the scalability issues of GPs to large data-sets. It reduces the computational cost to $O(m^2n)$, otherwise the cost of prediction would be $O(n^3)$, primarily influenced by the number of inducing points rather than the total dataset size. Furthermore, the stochastic optimization approach, utilizing mini-batches of data, facilitates efficient training on large datasets by minimizing the computational load per iteration and introducing regularization effects that aid in avoiding local minima. This methodology aligns with our objective to generate confidence scores associated with each object detection, where the ability to process extensive data efficiently and provide reliable measures of uncertainty is paramount. These features are also given to the RBF kernel for multi-class classification. The kernel is constructed from deep neural network features so that the parametric model can be integrated with the non-parametric model (GP). The Adam optimization algorithm \cite{kingma2014adam} is used to optimize the hyper-parameters of the GP model.

we utilise the joint optimization of the model parameters, including the deep neural network parameters and the GP hyperparameters, to achieve a cohesive and optimized model.

\subsection{End-to-End training of YOLOv3 Uncertainty-Aware Object Detection Network}

We train YOLOv3 in two stages. In the first stage, we train the objectness-detection and localization part by using data that has been automatically annotated by our 3D detector, as described in section \ref{Method_3DObjectness_Detection}. In the second stage, we train YOLO's object category recognition capability, coupled with uncertainty awareness capability, via a knowledge distillation method, using DCNN-GP as a ``teacher'' for the YOLO ``student'', as described in section \ref{Method_ObjectCategoryClassification}.

\begin{figure*}[ht]
%\centering
\begin{tabular}{c}
\subfloat{\includegraphics[width = 1.45in]{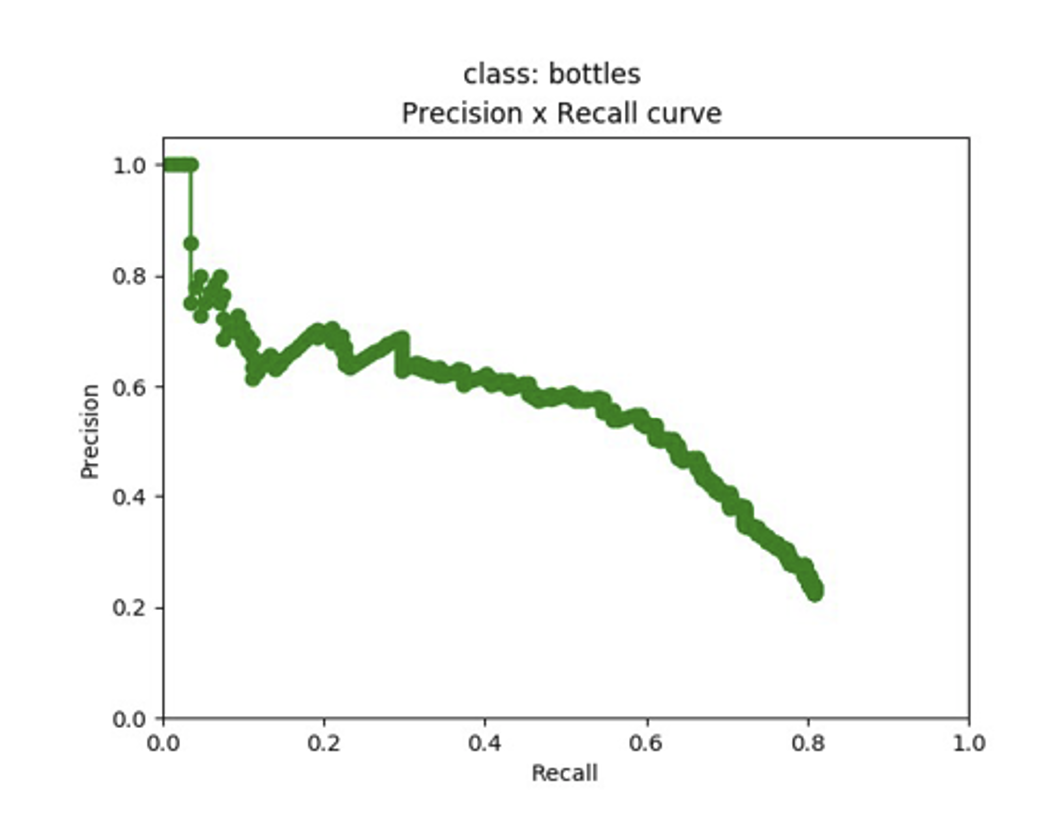}}
\subfloat{\includegraphics[width = 1.45in]{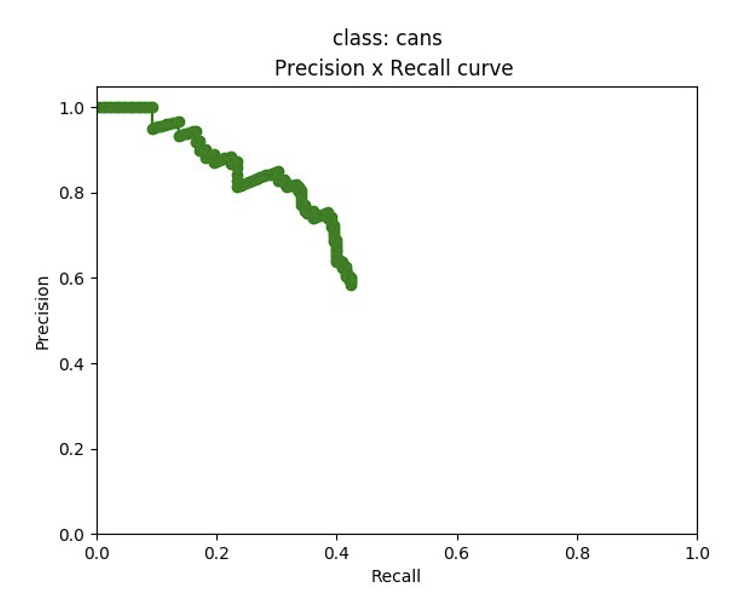}}
\subfloat{\includegraphics[width =1.45in]{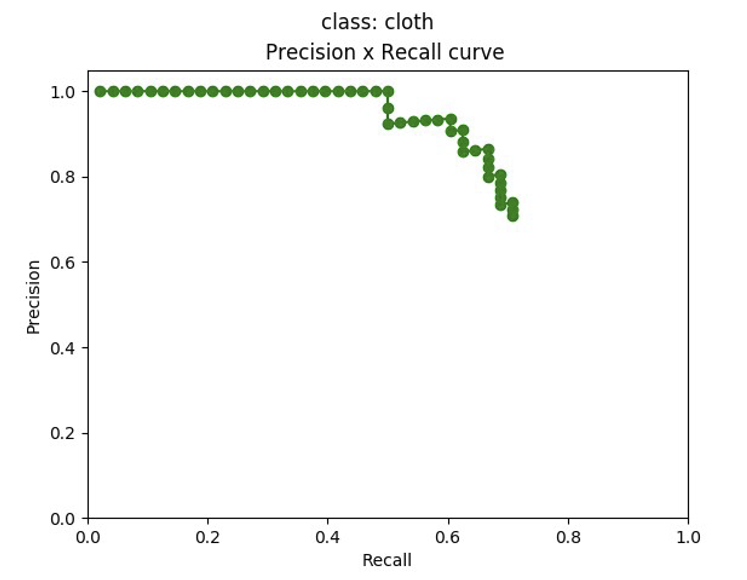}}
\subfloat{\includegraphics[width =1.45in]{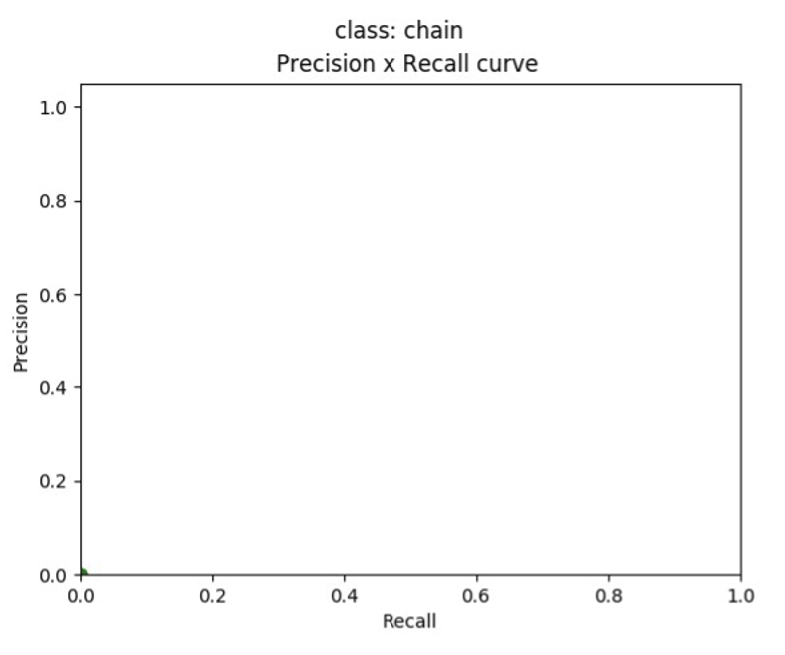}}
\subfloat{\includegraphics[width =1.45in]{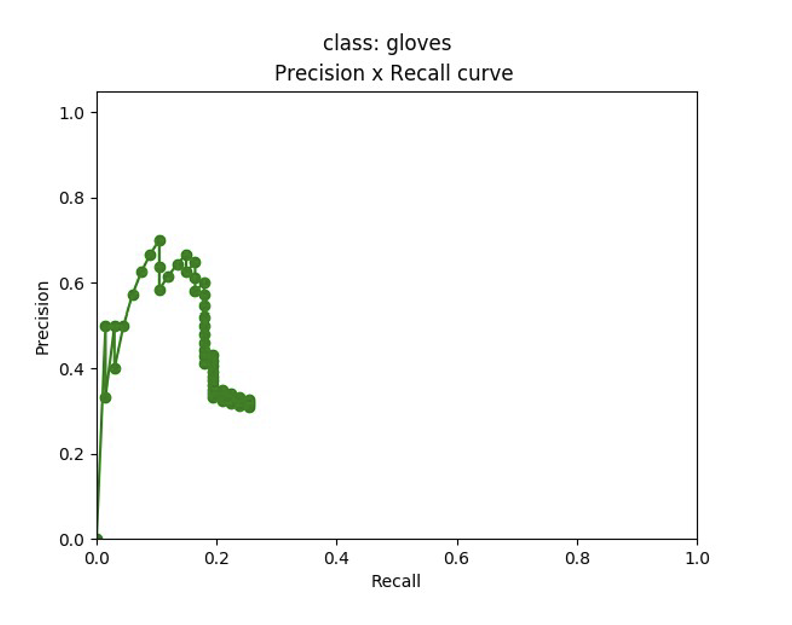}}\\
\subfloat{\includegraphics[width = 1.45in]{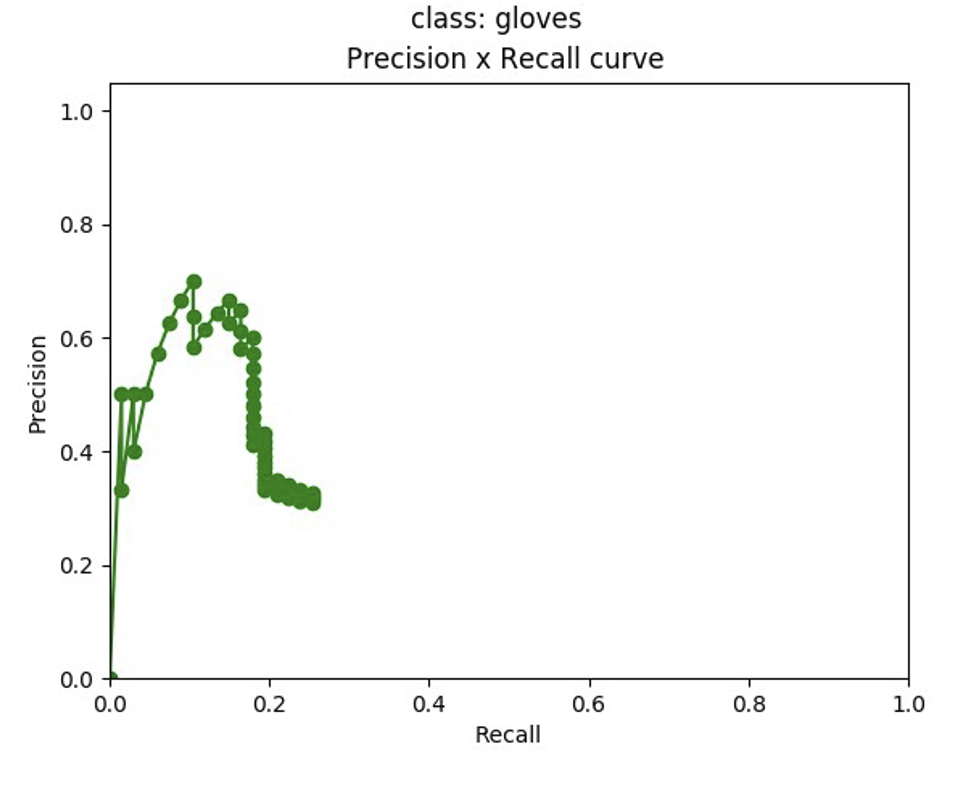}}
\subfloat{\includegraphics[width =1.45in]{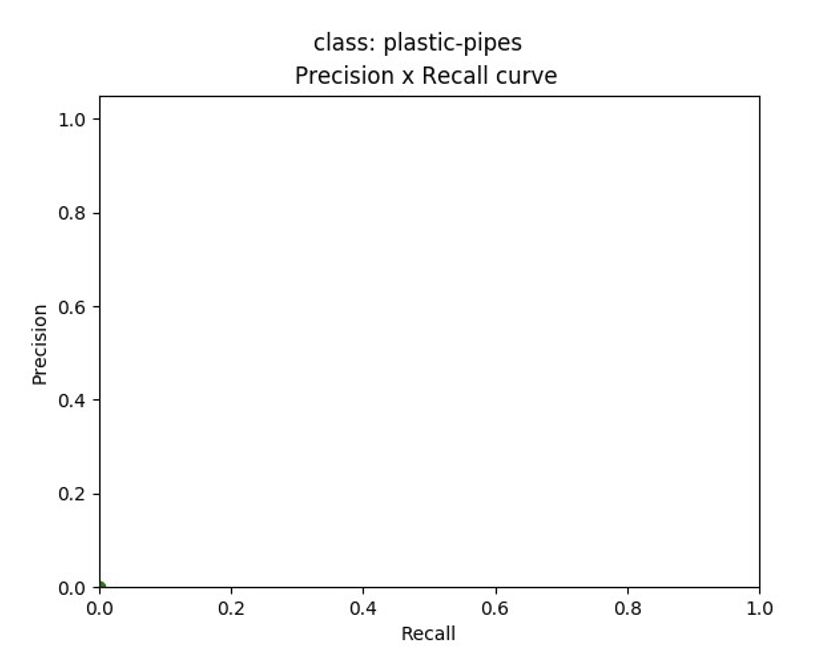}}
\subfloat{\includegraphics[width =1.45in]{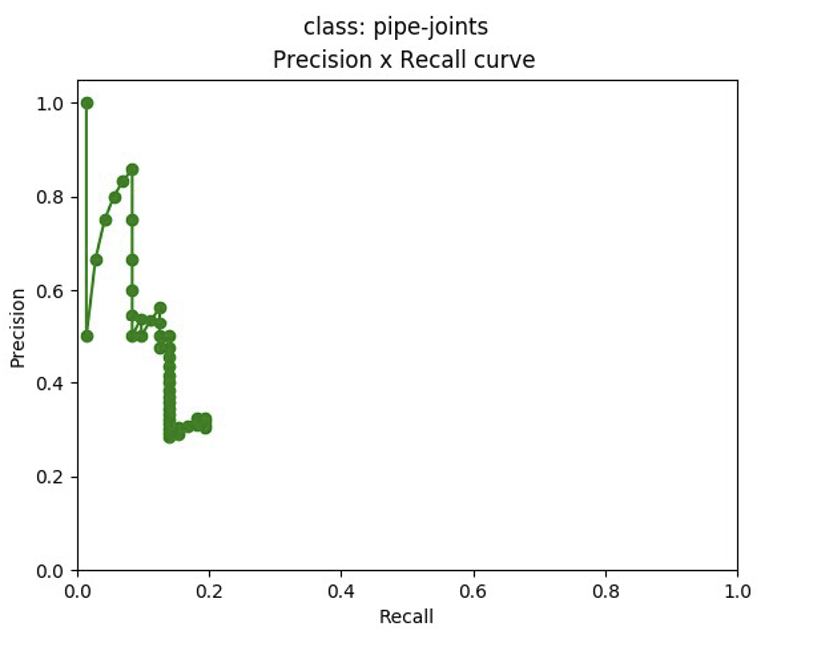}}
\subfloat{\includegraphics[width =1.45in]{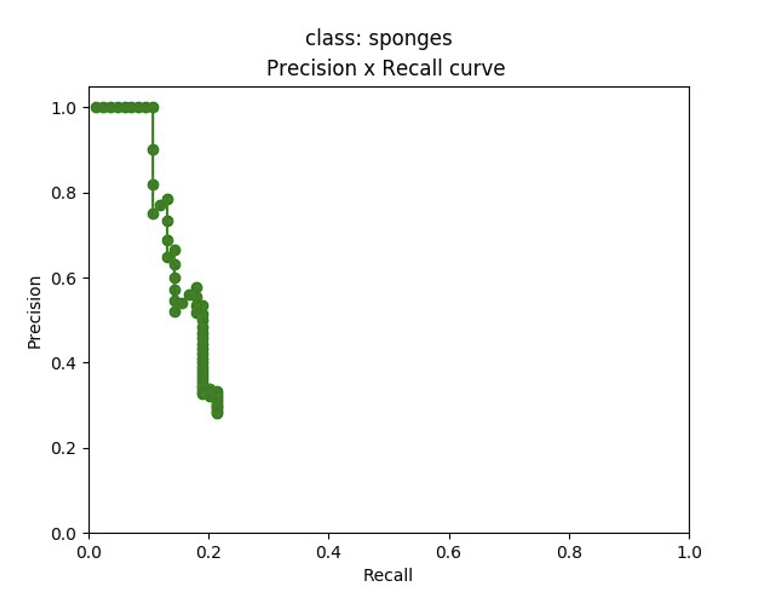}}
\subfloat{\includegraphics[width =1.45in]{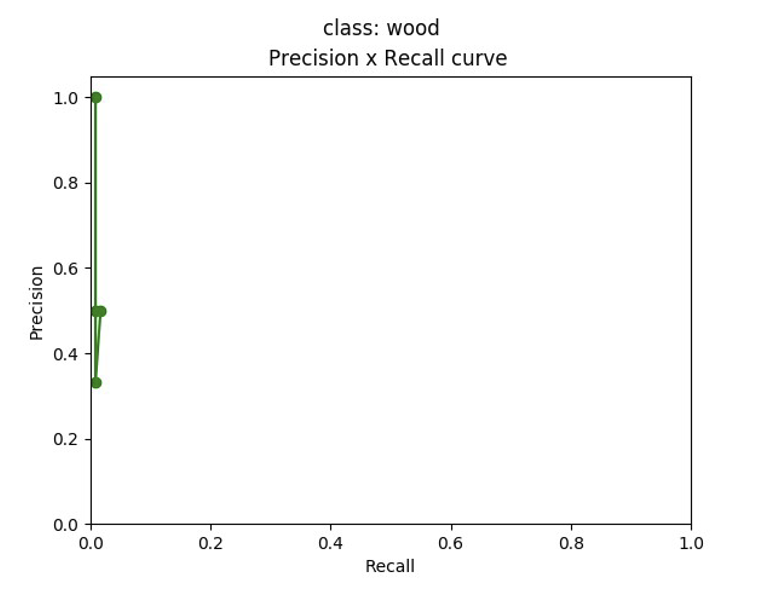}}
\end{tabular}

\begin{tabular}{c}
\subfloat{\includegraphics[width = 1.45in]{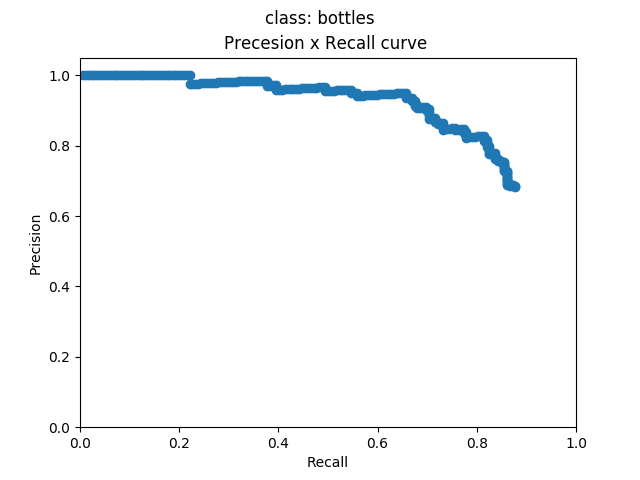}} 
\subfloat{\includegraphics[width = 1.45in]{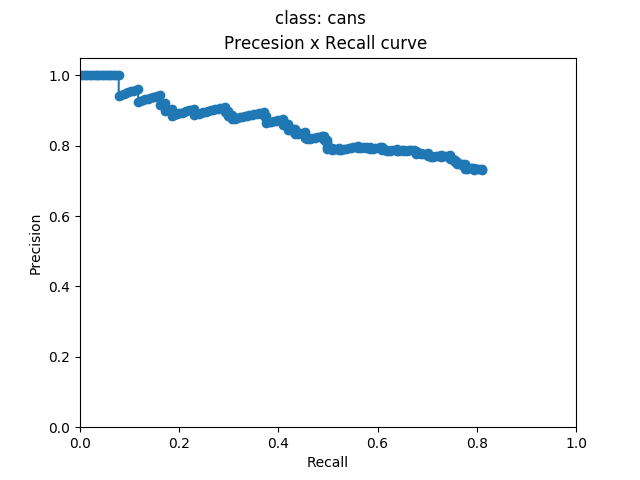}}
\subfloat{\includegraphics[width = 1.45in]{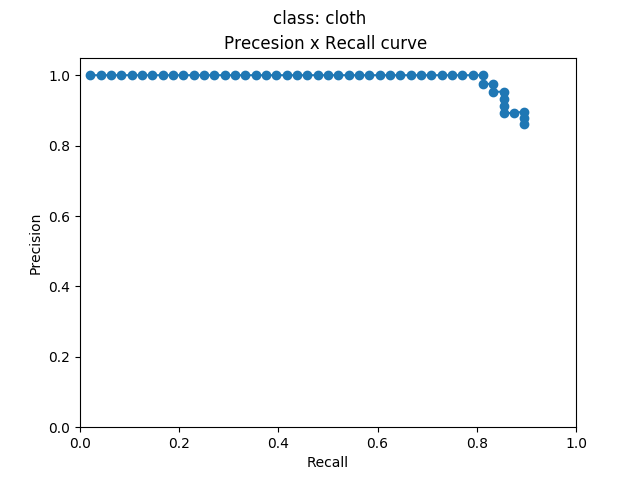}}
\subfloat{\includegraphics[width = 1.45in]{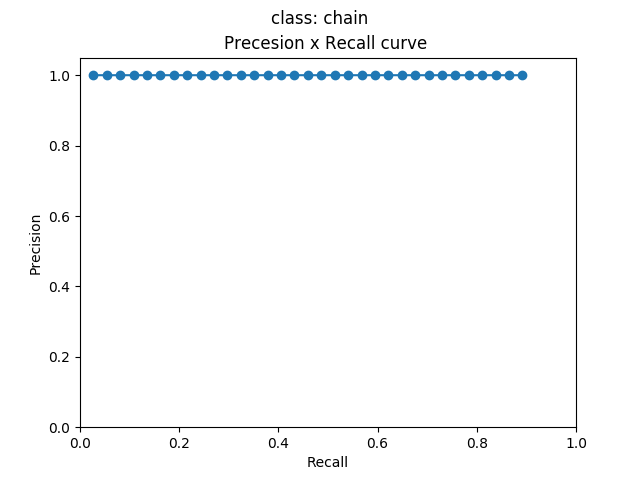}}
\subfloat{\includegraphics[width = 1.45in]{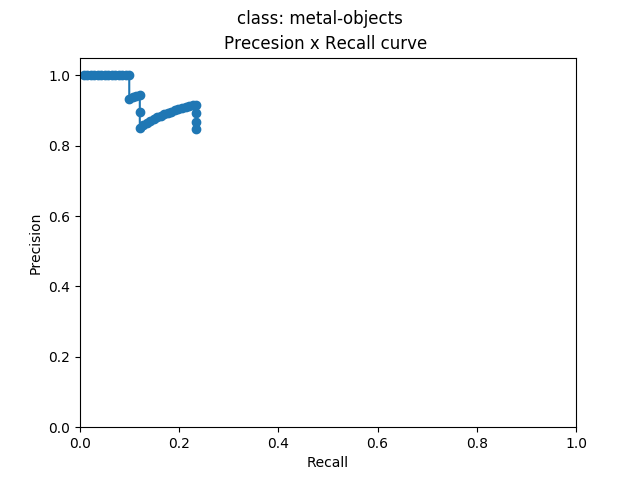}}\\
\subfloat{\includegraphics[width = 1.45in]{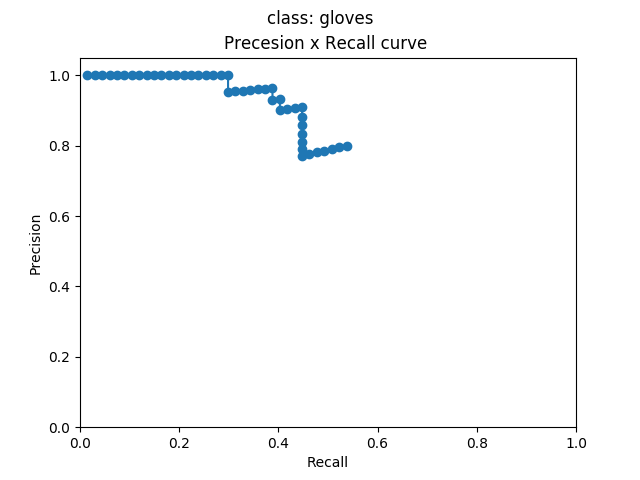}}
\subfloat{\includegraphics[width = 1.45in]{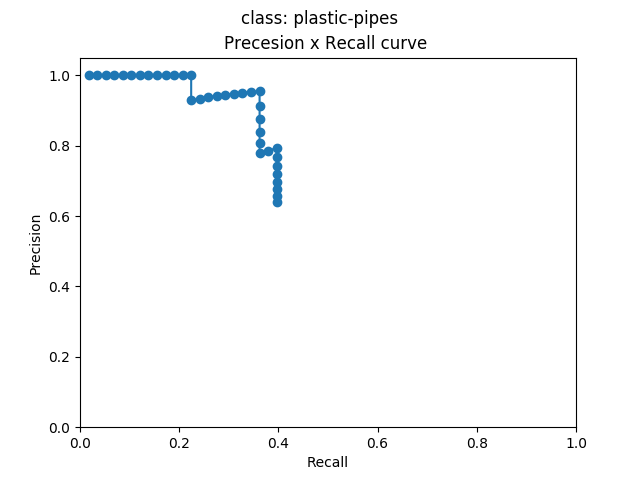}}
\subfloat{\includegraphics[width = 1.45in]{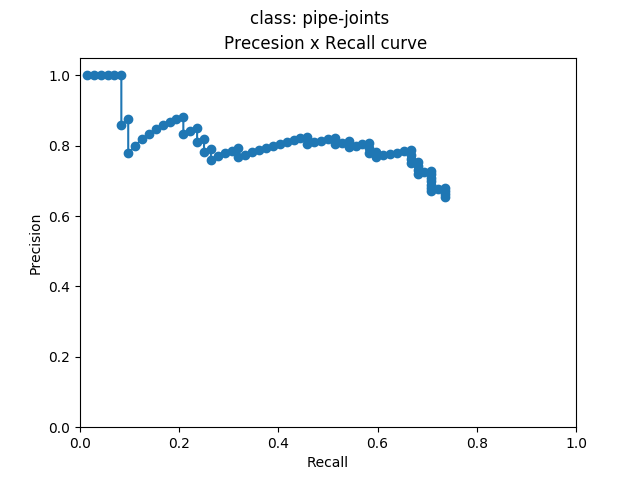}}
\subfloat{\includegraphics[width = 1.45in]{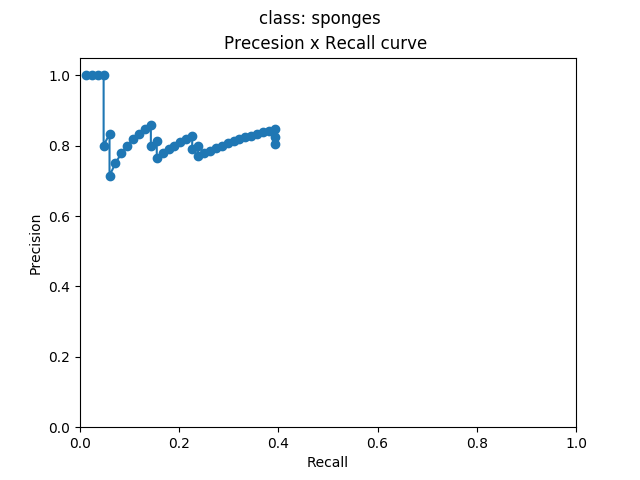}} 
\subfloat{\includegraphics[width = 1.45in]{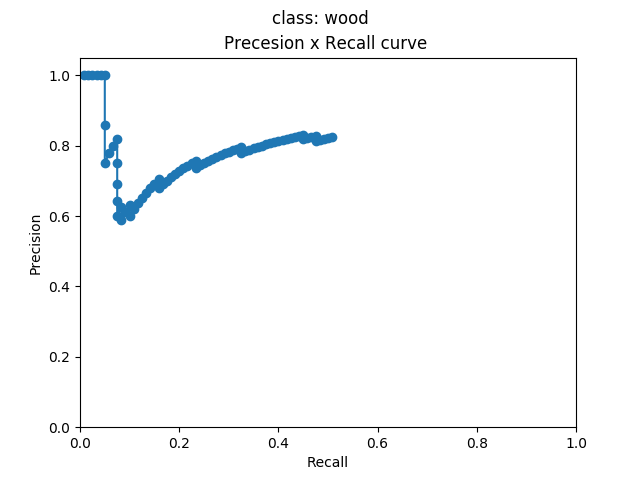}}

\end{tabular}
\caption{Precision-Recall curves, for each object category, of our nuclear waste test dataset for: i) conventional YOLOv3 (green curves, top); and ii) YOLOv3 trained using our knowledge distillation method (blue curves, bottom) trained using same amount of labelled data. It can be seen that our method achieves significantly better Precision-Recall trends on all object categories (i.e. false positives versus false negatives trade-off), highlighting superior performance in categorization decisions across various recall levels. Our method performs particularly strongly in categories where the conventional training method fails, e.g. 'bottles' and 'chains' with high precision.  While there are observed fluctuations in precision for 'metal-objects' and 'sponges', and a decline for some categories like 'cans' and 'cloth' at higher recall levels, overall performance is consistently good compared to conventional methods. Compared to the state-of-the-art YOLOv3, trained in conventional way on same data, our model maintains a commendable level of accuracy, particularly in challenging categories, reflecting the effectiveness of our algorithm and its potential as a benchmark for future enhancements in object detection.}
\label{fig:data}
\end{figure*}

In the conventional YOLOv3, the training phase utilizes the sum of squared error (SSE) loss for optimizing bounding box predictions and binary cross-entropy (BCE) loss for refining confidence scores and classifying detected objects.  

In contrast, we redesign the loss function for the classification task using the concept of knowledge distillation. The GP of the DCNN-GP ``teacher'' network, described in section \ref{Method_ObjectCategoryClassification}, is used to teach the YOLOv3 ``student' network how to predict confidence values for each categorization decision. We therefore use GP predictive class probabilities as input to YOLOv3, giving:

\begin{equation}
    \mathbf{Loss} = \sum_{i=1}^{n} (\sigma(\frac{\hat{y}^T}{T}) -\sigma(\frac{\hat{y}^S}{T}))^2
\end{equation}

where $\sigma$ refers to the Softmax function applied to logits and labels, labels are the predictive probabilities of our GPC model. $\textit{T}$ is utilized to smooth the outputs from the teacher classifier. Then, we apply the sum of squared error (SSE) on the logits and labels.

%%%%%%%%%%%%%%%%%%%%%%%%%%%%%%%%%%%%%%%%%%%%%%%%%%%%%%%%%%%%%%%%%%%%%%%%%%%%
%%%%%%%%%%%%%%%%%%%%%%%%%%%%%%%%%%%%%%%%%%%%%%%%%%%%%%%%%%%%%%%%%%%%%%%%%%%%%

\section{IMPLEMENTATION AND EXPERIMNENTS}

\subsection{Training through transfer learning with DCNN} \label{subsection:DCNN experiments}

In this study, we employed the ResNet50 model, which was pre-trained on the ImageNet dataset, to analyse a unique dataset consisting of nuclear waste-related objects. The pretrained weights of ResNet50 served as the foundational learning parameters for our specialized dataset, comprising 3,000 manually labelled cropped images depicting various forms of nuclear waste.
Our training and testing datasets were generated from rosbag files, created using a technique referenced in the work of [23] in the ROS (Robot Operating System) package [98]. Each class in the training set was represented by 300 randomly selected instances, while the test dataset comprised 100 instances per class previously unseen by the network. During the training, image scenes were cropped into smaller images and fed into the network with their respective class labels.
Our dataset is characterized by 10 distinct categories, including bottles, cans, chains, gloves, wood, pipe-joints, plastic-pipes, swabs (cleaning cloths), metal objects, and sponges. These items represent a spectrum of contaminated objects typically found in mixed intermediate-level (medium intensity gamma radiation contaminants) nuclear waste environments. Due to their deformable shapes (chains, rubber gloves) and reflective surfaces (metal objects), these objects pose a significant challenge for accurate classification.

For the training process using pre-trained ResNet50 through transfer learning in DCNN step, we utilised the Adam optimizer with a batch size of 32. We initiated the training with a learning rate of 0.001, reducing it by a factor of ten after 10,000 iterations, while applying a decay rate of 0.95. Notably, the training process converged after 8,000 iterations. 
As part of our methodology, we extracted features from ResNet50 and subsequently inputted these features into a Gaussian Process Classifier (GPC) model. This approach was necessary because the dataset exhibited complexities, particularly in distinguishing between similar items, such as plastic-pipes and pipe joints or wood and metal objects. As shown in Fig:1, the classifier can become confused
with some features of plastic-pipes with pipe joints and
similarly wood features with metal features.
As the project progressed, we observed a need to consider the background as a distinct class category. This adjustment was based on insights from the YOLOv3 detection system, which inherently classifies the background as one of its object categories. This observation was crucial for improving the accuracy of our object detection system, particularly in the challenging context of nuclear waste classification.

Our DCNN is based on PyTorch computational software\cite{NEURIPS2019_9015} and we use a high computational computer with i7-8 cores CPU a NVIDIA GTX 1070 for DCNN-GPC task and use NVIDIA TITAN RTX GPU 24GB for YOLOv3 training. During the transfer learning phase of our project, we utilized the ResNet50V2 network as a foundational model. To enhance efficiency and reduce training time, we kept the layers of ResNet50V2 frozen and integrated our layers specifically designed to extract features relevant to our dataset. This approach allowed us to minimize the computational resources and time required for training.

\subsection{Training uncertainty-awareness using GP model}

We teach uncertainty-awareness in two steps. The first step was explained in Subsection \ref{subsection:DCNN experiments}, where Resnet50 architecture is employed for feature extraction to transform raw input into a set of informative features for GP modelling. In the second step, we utilise a similar strategy and use the extracted features as input to build the kernel of the GP. We assign a distinct Gaussian Process to each feature emanating from the ResNet50 network. This one-to-one mapping between GPs and features allows for a fine-grained analysis and modelling of the intricate relationships inherent in the data.

We use 64 inducing points to estimate the variational distribution, which serves as an approximation of the prior distribution for the entire dataset. To elaborate, we transfer all the weight parameters from the initial stage of the pre-trained model directly into the deep Gaussian Process (GP) model. The model is trained for 100 epochs and an initial learning rate of $10^{-3}$ with an exponential decay of 0.95 is used. Here, Adam optimiser is used and 2 fully connected layers were added to the deep learning part of the network.

\subsection{End-to-end YOLOv3 training}

We propose a modified approach wherein YOLOv3 is trained using knowledge distillation techniques, adopting the role of a 'student' classifier. This involves integrating feature inputs from a Gaussian Process Classifier (GPC) model into YOLOv3. Our strategy diverges from the conventional loss function outlined in \cite{hinton2015distilling}, which we found unsuitable for our dataset characterized by a high potential for misclassification across multiple object categories as shown in Fig.\ref{fig:testImgYolo} Instead, we have re-engineered the loss function to smooth the logits of the teacher model effectively.

A key aspect of knowledge distillation is the manipulation of the ``Temperature'' parameter in the teacher model to yield softer probability distributions, thereby reducing overconfidence in predictions. Given the complex nature of the features in our dataset, with difficult to distinguish objects appearing with diverse shapes, we empirically determined a Temperature value of 2 to be optimal. This adjustment ensures the generation of softened probabilities and more evenly distributed prediction outcomes without sharp peaks.

Furthermore, our experimental analysis revealed that the traditional loss metrics employed in knowledge distillation, such as Kullback-Leibler (KL) divergence, Mean Squared Error (MSE), Mean Absolute Error (MAE), and Cross-Entropy Loss, were ineffective for our dataset. Instead, we found that Sum of Squared Errors (SSE) loss, particularly when combined with a calibrated temperature setting, yielded significantly better results, aligning closely with the complex feature profiles of the objects in our dataset.

\begin{figure}[h]
    \centering
    \includegraphics[width=\columnwidth]{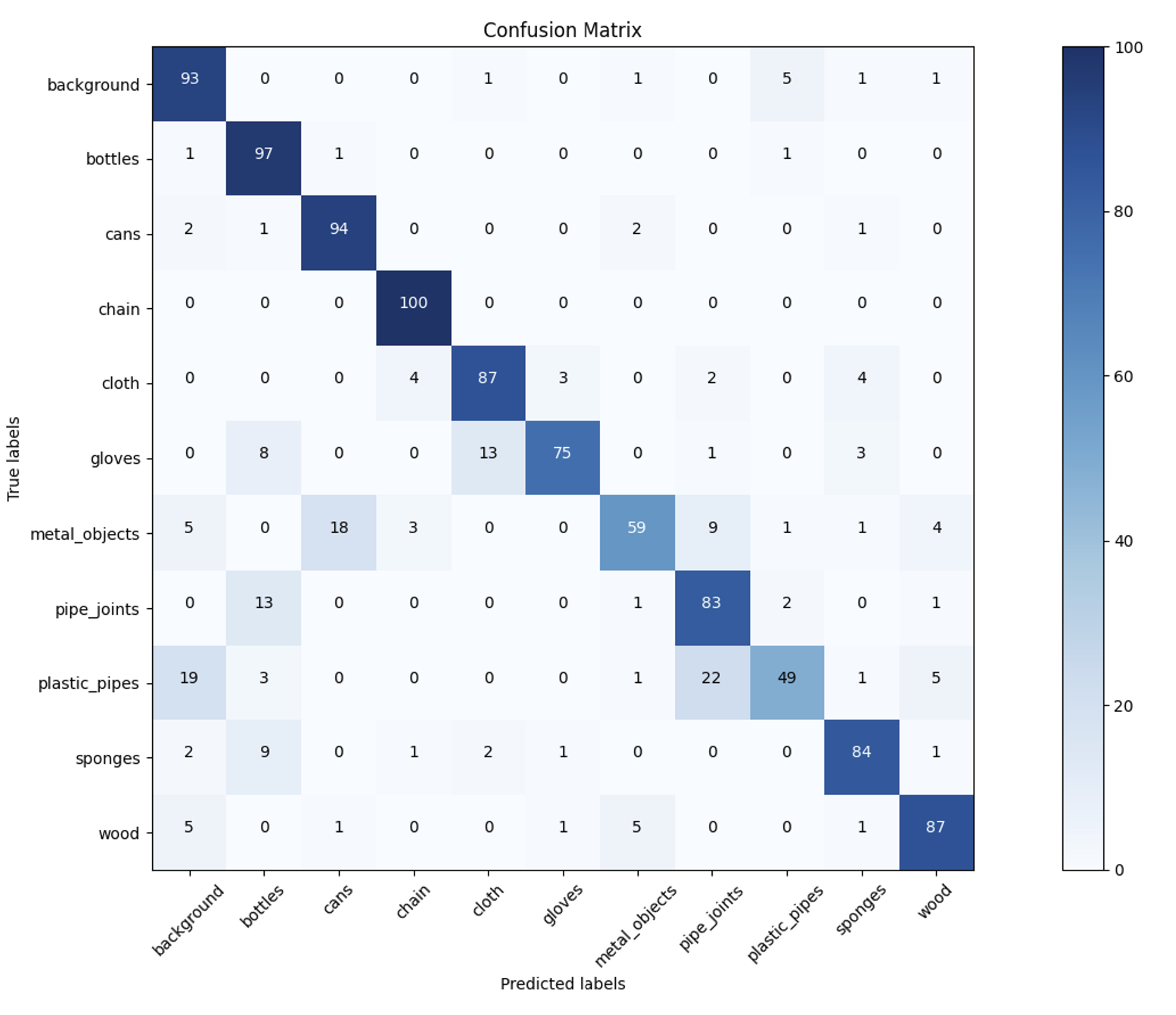}
    \caption{Confusion matrix DCNN-GP: representing the performance of the overall DCNN-GP model. The row represents the actual classes and the column represents the predicted classes.  The confusion matrix depicts the number of True Positives (TP), True Negatives (TN), False Positives (FP), and False Negatives (FN) predictions made by the model on the test dataset. Our model achieved a Precision of 85.4, a Recall of 81.0, and an F-Score of 82.0, indicating a robust balance between precision and recall. Notably, the low number of FN and FP demonstrates the model's efficacy in minimizing Type I and Type II errors.    }
    \label{fig:Confusion_matrix}
\end{figure}

\section{EVALUATION}

\begin{table*}[t]
  \centering
  \caption{Precision, Recall and F-Score on all object categories for: baseline method \cite{Kevinwork}; a YOLOv3 network trained in conventional way using the same amount of labelled data; our proposed method of YOLOv3 trained using teacher-student paradigm with knowledge distillation.}

  \label{acc}
  \begin{tabular}{c |c|c|c}
    \hline
    \cellcolor{gray!30}Category & \cellcolor{green!30}\textbf{Baseline method} & \cellcolor{red!30}\textbf{Conventional YOLOv3} & \cellcolor{purple!55}\textbf{Our method} \\
    \hline \hline
    & \begin{tabular}{c|c|c}
          \textbf{Precision} & \textbf{Recall} & \textbf{F-Score}
      \end{tabular} & 
      \begin{tabular}{c|c|c}
          \textbf{Precision} & \textbf{Recall} & \textbf{F-Score}
      \end{tabular} & 
      \begin{tabular}{c|c|c}
          \textbf{Precision} & \textbf{Recall} & \textbf{F-Score}
      \end{tabular} \\
    \hline \hline
    bottles & \begin{tabular}{c c c}
        89.2 & 83.2 & 86.1
      \end{tabular} & 
      \begin{tabular}{c c c}
        40.0 & 69.0 & 51.0
      \end{tabular} & 
      \begin{tabular}{c c c}
        83.0 & \textbf{98.0} & \textbf{90.0}
      \end{tabular} \\
    \hline
    cans & \begin{tabular}{c c c}
        81.8 & 91.8 & 86.6
      \end{tabular} & 
      \begin{tabular}{c c c}
        74.1 & 45.0 & 56.0
      \end{tabular} & 
      \begin{tabular}{c c c}
        79.0 & \textbf{92.0} & 85.0
      \end{tabular} \\
    \hline
    chains & \begin{tabular}{c c c}
        79.2 & 95.0 & 86.4
      \end{tabular} & 
      \begin{tabular}{c c c}
        0 & 0 & 0
      \end{tabular} & 
      \begin{tabular}{c c c}
        \textbf{100.0} & 95.0 & \textbf{97.0}
      \end{tabular} \\
    \hline
    cloth & \begin{tabular}{c c c}
        93.3 & 80.0 & 86.2
      \end{tabular} & 
      \begin{tabular}{c c c}
        89.4 & 70.8 & 79.0
      \end{tabular} & 
      \begin{tabular}{c c c}
        91.0 & \textbf{96.0} & \textbf{93.0}
      \end{tabular} \\
    \hline
    gloves & \begin{tabular}{c c c}
        68.3 & 91.5 & 78.2
      \end{tabular} & 
      \begin{tabular}{c c c}
        62.0 & 26.8 & 37.4
      \end{tabular} & 
      \begin{tabular}{c c c}
        \textbf{83.0} & 66.0 & 73.0
      \end{tabular} \\
    \hline
    metal-objects & \begin{tabular}{c c c}
        75.0 & 64.0 & 69.1
      \end{tabular} & 
      \begin{tabular}{c c c}
        50.0 & 0.7 & 1.4
      \end{tabular} & 
      \begin{tabular}{c c c}
        \textbf{93.0} & 53.0 & 68.0
      \end{tabular} \\
    \hline
    pipe-joints & \begin{tabular}{c c c}
        66.7 & 90.2 & 76.7
      \end{tabular} & 
      \begin{tabular}{c c c}
        57.7 & 20.8 & 30.6
      \end{tabular} & 
      \begin{tabular}{c c c}
        \textbf{72.0} & \textbf{92.0} & \textbf{81.0}
      \end{tabular} \\
    \hline
    plastic-pipes & \begin{tabular}{c c c}
        63.2 & 50.0 & 55.8
      \end{tabular} & 
      \begin{tabular}{c c c}
        0 & 0 & 0
      \end{tabular} & 
      \begin{tabular}{c c c}
        \textbf{79.0} & \textbf{64.0} & \textbf{71.0}
      \end{tabular} \\
    \hline
    sponges & \begin{tabular}{c c c}
        92.5 & 87.5 & 89.9
      \end{tabular} & 
      \begin{tabular}{c c c}
        47.3 & 22.7 & 30.7
      \end{tabular} & 
      \begin{tabular}{c c c}
        87.0 & 82.0 & 84.0
      \end{tabular} \\
    \hline
    wood & \begin{tabular}{c c c}
        87.8 & 87.8 & 87.8
      \end{tabular} & 
      \begin{tabular}{c c c}
        1.0 & 2.5 & 2.8
      \end{tabular} & 
      \begin{tabular}{c c c}
        87.0 & 70.0 & 78.0
      \end{tabular} \\
    \hline
    \cellcolor{yellow!70}Overall Avg. & \begin{tabular}{c c c}
        80.9 & 83.5 & 82.2
      \end{tabular} & 
      \begin{tabular}{c c c}
        49.36 & 18.83 & 28.89
      \end{tabular} & 
      \begin{tabular}{c c c}
        \textbf{85.4} & 81.0 & 82.0
      \end{tabular} \\
    \hline
  \end{tabular}
  \label{table:comparison}
\end{table*}

Table \ref{table:comparison} shows the Precision, Recall and F-Score values for each object category, for: our method (YOLOv3 trained with Knowledge Distillation); a conventionally trained YOLOv3 network using the same amount of labelled training data; and the baseline method \cite{Kevinwork}. Figure \ref{fig:Confusion_matrix} shows the confusion matrix for our method across all object categories. These results demonstrate the consistent and competitive performance of our object detection method compared to both the baseline method \cite{Kevinwork} and conventional YOLOv3 across various categories in a realistic mockup nuclear waste object dataset. Notably, our method yielded an overall precision of 85.4\%,
which is a significant achievement considering the complexity and challenges presented by our dataset.

Note that comparison with the ``baseline'' method is not strictly fair, because\cite{Kevinwork} makes use of full 3D depth data for classification, whereas (after training) our YOLOv3 method is a 2D classifier and only makes use of 2D RGB data. However we include the comparison data for completeness, since \cite{Kevinwork} is the literature most closely related to our work, and is tested on the same object dataset.

From the table, it can be seen that our approach achieves markedly higher precision, recall, and F-score compared with conventional YOLOv3. It also significantly outperforms the baseline method, especially on categories where the baseline method was weakest, while still maintaining competitive scores on categories where the baseline performed strongly. Our method shows significant improvements on `bottles', `chains', `cloth', `pipe-joints' and `plastic-pipes' categories, indicating robust feature recognition and classification capabilities. The consistently strong F-scores across most categories underscore the effectiveness of our method in accurately detecting and classifying diverse objects. Even though \cite{Kevinwork} uses a 3D-based detector, it performs less well than our method on shiny objects such as metal objects and pipes. This is likely because of missing depth data on shiny objects which can confuse point-cloud cameras.

 In comparison to the baseline method documented by \cite{Kevinwork}, our approach not only demonstrates improved Precision but also comparable Recall and F-score metrics. Moreover, our method exhibits a faster execution time of 40-45ms per image for detection tasks, significantly outpacing the baseline's 100-200ms. This improvement in computational efficiency makes our method more useful for real-time applications such as vision-guided robotic waste handling \cite{OECD2022}, and potentially more scalable to more extensive or demanding tasks in future applications.

 Our proposed method also provides an additional functionality, since the baseline method of \cite{Kevinwork} is not capable of outputting confidence/uncertainty scores alongside each of its categorizations.

%%%%%%%%%%%%%%%%%%%%%%%%%%%%%%%%%%%%%%%%%%%%%%%%%%%%%%%%%%%%%%%%%%%%%%%%%%%%%%%%%%%%%
%%%%%%%%%%%%%%%%%%%%%%%%%%%%%%%%%%%%%%%%%%%%%%%%%%%%%%%%%%%%%%%%%%%%%%%%%%%%%%%%%%%%%

\section{Conclusion}
This paper has presented a novel knowledge distillation paradigm for ``self-supervised'' learning, requiring minimal labelled training data. This is particularly useful for many industrial applications where large annotated datasets are often unavailable.

In contrast to conventional knowledge distillation methods (which ``distil'' the knowledge of a complex network into a simpler classifier), our approach uses a weakly-trained ``teacher'' classifier to automatically annotate additional training data, to train a more powerful ``student'' network based on a modified YOLOv3 network.

We utilize the predictive probabilities from a hybrid model combining DCNN and GP models, to enhance the performance of the YOLOv3 object detector. This enables the modified YOLOv3 network to output significantly improved confidence scores alongside each classification.  By incorporating uncertainty levels into the perception task, the proposed system can facilitate robot decision-making with increased accuracy. Such uncertainty-aware decision-making is particularly important in safety-critical industries like nuclear decommissioning or handling other hazardous waste forms such as, e.g. disassembly of the large (and potentially flammable or explosive) lithium-ion battery packs of electric vehicles for recycling and circular economy \cite{rastegarpanah2021towards, harper2019recycling}.

Moving forward, future research will explore the application of the proposed ideas in this paper to 3D data-sets, while simultaneously incorporating uncertainties in both localization and object categorization into the YOLOv3 or its subsequent versions like YOLOX. Additionally, we are exploring the combination of these object localisation and categorization algorithms with our lab's work on advanced robotics methods for vision-guided autonomous grasping and manipulation \cite{adjigble2018model, adjigble2021spectgrasp, adjigble2023haptic}.

\section*{Acknowledegment}

This work was supported by the REBELION project 101104241, UK Industrial Strategy Challenge Fund (ISCF) EP/R02572X/1, CHIST-ERA EP/S032428/1, and UKRI EP/P01366X/1, EP/P017487/1.

\section*{Data and code Availability statement}
The data produced or examined in this study and the code utilized for the analysis can be requested from the author.

\bibliographystyle{IEEEtran}

\bibliography{Biblography.bib}

\end{document}